\documentclass[article,10pt]{IEEEtran}

\usepackage[utf8]{inputenc} 
\usepackage[T1]{fontenc}
\usepackage{url}              % provides \url{...}
\usepackage{cite}             % improves presentation of citations

\usepackage[cmex10]{amsmath} 
\usepackage{mleftright}      
\mleftright                  
\usepackage{amsmath}
\DeclareMathOperator*{\argmin}{argmin}
\usepackage{graphicx}         
\usepackage{booktabs}         
\usepackage{mathtools}
     
\usepackage{graphicx,cite,amsmath,amssymb,amsfonts,mathtools,algorithm,mathabx,textcomp,blindtext,xcolor,subfig,float,mwe,dblfloatfix,bm, xfrac,algpseudocode, stackengine, verbatim,xparse} 
\usepackage[font=normalsize]{caption}
\usepackage{amsthm}
\newtheorem{remark}{Remark}
\newtheorem{assumption}{Assumption}
\newtheorem{proposition}{Proposition}
\newtheorem{corollary}{Corollary}

\begin{document}
\title{Partially Decentralized Multi-Agent Q-Learning via Digital Cousins for Wireless Networks
	\thanks{Talha Bozkus and Urbashi Mitra are with the Ming Hsieh Department of Electrical and Computer Engineering, University of Southern California, Los Angeles, USA. Email: \{bozkus, ubli\}@usc.edu. }
	\thanks{This work was funded by the following grants:  ARO W911NF1910269, ARO W911NF2410094,  NSF CIF-2311653, NSF CIF-2148313, NSF RINGS 2148313,  ONR N00014-22-1-2363, NSF DBI-2412522, and is also supported in part by funds from federal agency and industry partners as specified in the Resilient \& Intelligent NextG Systems (RINGS) program).}
}

\author{Talha Bozkus and Urbashi Mitra}

\maketitle

\begin{abstract}
$Q$-learning is a widely used reinforcement learning (RL) algorithm for optimizing wireless networks, but faces challenges with large state-spaces. Recently proposed multi-environment mixed $Q$-learning (MEMQ) algorithm addresses these challenges by employing multiple $Q$-learning algorithms across multiple synthetically generated, distinct but structurally related environments, so-called {\em digital cousins}. In this paper, we propose a novel multi-agent MEMQ (M-MEMQ) for \emph{cooperative decentralized} wireless networks with multiple networked transmitters (TXs) and base stations (BSs). TXs do not have access to global information (joint state and actions). The new concept of  \emph{coordinated} and \emph{uncoordinated} states is introduced. In uncoordinated states, TXs act independently to minimize their individual costs and update local $Q$-functions. In coordinated states, TXs use a Bayesian approach to estimate the joint state and update the joint $Q$-functions. The cost of information-sharing scales linearly with the number of TXs and is independent of the joint state-action space size. Several theoretical guarantees, including deterministic and probabilistic convergence, bounds on estimation error variance, and the probability of misdetecting the joint states, are given. Numerical simulations show that M-MEMQ outperforms several decentralized and centralized training with decentralized execution (CTDE) multi-agent RL algorithms by achieving 60\% lower average policy error (APE), 40\% faster convergence, 45\% reduced runtime complexity, and 40\% less sample complexity. Furthermore, M-MEMQ achieves comparable APE with significantly lower complexity than centralized methods. Simulations validate the theoretical analyses.
\end{abstract}


\section{Introduction}

Reinforcement learning (RL) is a powerful framework for optimizing next-generation wireless networks that can be modeled as Markov Decision Processes (MDPs) \cite{q_wireless_1, q_wireless_2}. When the MDP dynamics, such as transition probabilities and cost functions, are unknown, model-free RL algorithms such as $Q$-learning can be employed. $Q$-learning is a simple, tabular, and model-free RL algorithm and works with discrete and modest-sized state-action spaces \cite{q_learning_survey}. However, it faces several challenges, including slow convergence, high sample complexity, high estimation bias, and variance, and training instability across large state-action spaces \cite{pn_journal, ln_journal}.

We have proposed multi-environment mixed $Q$-learning (MEMQ) to tackle these issues by running multiple $Q$-learning algorithms across multiple, distinct yet structurally related synthetic environments -- \emph{digital cousins} and fusing their $Q$-functions into a single ensemble $Q$-function estimate using adaptive weighting \cite{pn_journal, ln_journal, talha_eusipco}. MEMQ has three distinct features compared to the existing $Q$-learning algorithms. First, MEMQ constructs multiple, synthetic, distinct yet structurally related environments during training. These environments have been shown to help accelerate the exploration of the state-space \cite{pn_journal}. Second, MEMQ trains multiple $Q$-learning agents on these environments and combines their outputs into an ensemble $Q$-function to achieve more robust $Q$-functions with lower variance \cite{pn_journal}. Third, MEMQ uses a mixed-learning strategy -- it combines real-time interactions with synthetic samples collected from the synthetic environments for faster training with improved sample complexity \cite{pn_journal, ln_journal}.

MEMQ was originally designed for single-agent centralized MISO and MIMO wireless networks to solve real-world challenges such as resource allocation and interference minimization with lower complexity. However, it is not directly applicable to multi-agent and decentralized wireless networks. To this end, multi-agent RL (MARL) theory can be used to extend single-agent MEMQ to multi-agent networks. MARL has been effectively applied to different wireless network optimization problems, including resource and power allocation, channel access, beamforming, spectrum management, and multi-access edge computing \cite{wireless_survey_1, wireless_survey_2, coordinated_State_5}, and can be categorized along several dimensions. Different training strategies depend on how agents are trained and execute their decisions and include fully centralized \cite{pn_journal, 
randomized_double_q}, centralized training decentralized execution (CTDE) \cite{ctde_1}, decentralized with networked agents (DNA) \cite{networked_agent_1, networked_agent_3} and fully decentralized \cite{independent_1, independent_2}. If agents have complete visibility of the global state, it is fully observable; otherwise partially observable \cite{based_on_problem_descrptipion}. Different communication strategies may vary in terms of which agents communicate, when and how to establish communication links, what information to share, and how to combine the received messages \cite{based_on_communication_strategies}. Different coordination strategies depend on whether agents collaborate to achieve a common goal and include fully cooperative \cite{cooperative_1}, fully competitive \cite{competitive_1}, or mixed \cite{mixed_cooperation_1}. If agents share the same observation and action space and have a common reward function, then they are \textit{homogeneous} \cite{homogeneous_agent_1}; otherwise \textit{heterogeneous} \cite{heterogenous_agent_1}. Other categorizations include value-based vs. policy-based methods, tabular vs. deep neural network approaches, online vs. offline vs. hybrid methods, and methods for discrete vs. continuous state-action spaces.

In this work, we develop a multi-agent MEMQ (M-MEMQ) algorithm for wireless networks with multiple decentralized but networked agents (i.e. transmitters). The system comprises multiple mobile transmitters (agents) and stationary base stations. The overall state-space is divided into \textit{coordinated} and \textit{uncoordinated} states. In uncoordinated states, agents operate under partial observability in a decentralized manner without access to global information (e.g., other agents' states, actions, or costs) and independently update their local $Q$-tables. In coordinated states, agents can locally measure the aggregated received signal strength (ARSS) due to other transmitters to estimate the joint state using maximum a posteriori probability (MAP) estimation and communicate with the leader agent by sharing limited information (i.e., their local $Q$-functions and costs) to cooperatively minimize the joint cost. Agents are homogeneous and share the same state-action spaces and cost functions. M-MEMQ inherits key characteristics of MEMQ -- it is a value-based, off-policy algorithm and adopts a mixed-learning strategy. It is also a tabular algorithm and thus is applicable to moderately large discrete state-action spaces. The tabular setting enables rigorous theoretical analysis, including convergence, stability, variance reduction, and sample-complexity guarantees. Moreover, many wireless networking problems can still be effectively represented using discretized state-action spaces \cite{q_learning_additional_ref_1, q_learning_additional_ref_2, menevse2025accelerated}. The proposed digital-cousin and multi-environment learning principles may also be integrated with deep RL methods for continuous-space systems, which is beyond the scope of this work.

MARL algorithms face challenges such as partial observability, scalability, effective exploration and security concerns, non-stationary, and credit assignment \cite{marl_challenges}. Our proposed M-MEMQ algorithm can alleviate these challenges effectively. For example, while global information is not available to agents, they use MAP estimation based on their local observations to estimate the joint state. On the other hand, M-MEMQ offers a good balance between policy accuracy, runtime complexity, and communication costs across an increasing number of agents and larger state-action space, making it scalable. M-MEMQ also accelerates the exploration of the joint state-space by combining real-time interactions with synthetic samples collected from synthetic environments in real-time. The problem of non-stationary is alleviated through the leader agent, which uses a centralized $Q$-function update by aggregating information from individual agents to maintain a global system view. Finally, the leader agent updates the global $Q$-functions and determines the joint state and action, considering each agent’s confidence in their joint state estimates, which prioritizes more confident agents by assigning them greater weight (i.e., credit) for their contributions.

Our multi-agent MEMQ algorithm introduces substantial improvements over the single-agent MEMQ framework \cite{pn_journal, ln_journal}. We employ a minimal-overhead communication protocol where agents share only local Q-values and costs, ensuring scalability linear in the number of agents rather than the joint state–action space \cite{based_on_cooperation_type}. To overcome partial observability, we use Bayesian joint-state estimation from local observations without requiring state information exchanges \cite{based_on_cooperation_type}. We systematically evaluate and optimize different digital cousin architectures (shared, common, independent) for decentralized environments. Our algorithm supports real-time adaptive coordination, dynamically switching between coordinated and uncoordinated modes using ARSS measurements instead of fixed assumptions \cite{coordinated_State_5, coordinated_State_8}. Finally, we design MEMQ-specific Q-function update rules tailored to the multi-environment setting, which differ fundamentally from prior MARL. We also introduce new theoretical contributions that are mostly missing in the partially decentralized MARL literature \cite{MARL_SURVEY_0, MARL_SURVEY_1}. We establish both probabilistic and deterministic convergence results and upper bounds on the variance of estimation errors for the joint $Q$-functions, and provide sample complexity guarantees for achieving accurate policies. We further analyze the criterion for distinguishing coordinated from uncoordinated states and quantify the probability of misdetection under noisy ARSS, whereas prior works treat this selection as fixed or heuristic without theoretical justifications \cite{coordinated_State_4, coordinated_State_5}.

{Our approach herein is a form of \textit{generative} learning, but differs from traditional generative methods such as generative adversarial networks \cite{generative_survey}. We construct synthetic environments -- synthetic simulators that RL agents can continuously interact with to generate synthetic training data alongside real data during training. Our approach is computationally efficient to implement, interpretable, and has strong theoretical guarantees. The synthetically generated data herein follow different, but related distribution to the real-time data and are augmented with real-time data to accelerate exploration \cite{pn_journal, ln_journal}.} We observe that due to our exemplar application, wireless communication networks, our distinct agents actually experience different environments and dynamics in contrast to many MARL works \cite{wireless_survey_1, wireless_survey_2}.

The \textbf{main contributions} of this paper are as follows: (i) We develop the first multi-agent MEMQ algorithm (M-MEMQ) for decentralized wireless networks with networked agents. (ii) In coordinated states, agents estimate the joint state using a Bayesian MAP estimation based on local observations in a decentralized way to minimize the cost of information exchange. In the uncoordinated states, agents act independently. (iii) We provide the first theoretical guarantees for multi-agent MEMQ algorithms, including deterministic and probabilistic convergence, bounds on estimation error variance, and the probability of misdetecting the joint states. (iv) We evaluate our algorithm on a wireless network with multiple mobile transmitters (TX) and stationary base stations (BS) to find the optimal TX-BS association to minimize the joint wireless cost. Our algorithm outperforms several decentralized and CTDE-based MARL algorithms by achieving 60\% lower average policy error (APE), 40\% faster convergence rate, 45\% less computational runtime complexity and 40\% less sample complexity. Our algorithm achieves comparable APE with significantly lower complexity than centralized methods. In addition, our algorithm exhibits high robustness to changes in network parameters. Simulations verify theoretical analyses.

We use the following notation: the vectors are bold lower case (\textbf{x}), matrices and tensors are bold upper case (\textbf{A}), and sets are in calligraphic font ({$\mathcal{S}$}).

\setlength{\textfloatsep}{5pt}
\begin{figure*}[h]
    \centering
    \subfloat[Q-learning]{{\includegraphics[width=2.2cm]{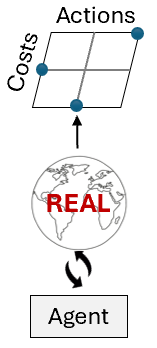}}}%
    \hspace{5pt}
    \subfloat[Single-agent MEMQ]{{\includegraphics[width=3.6cm]{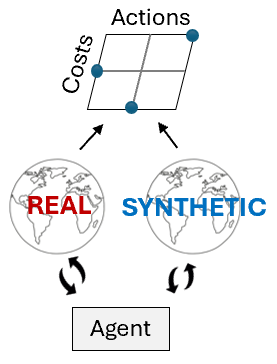}}}%
    \hspace{5pt}
    \subfloat[Proposed Multi-Agent MEMQ]{{\includegraphics[width=10cm]{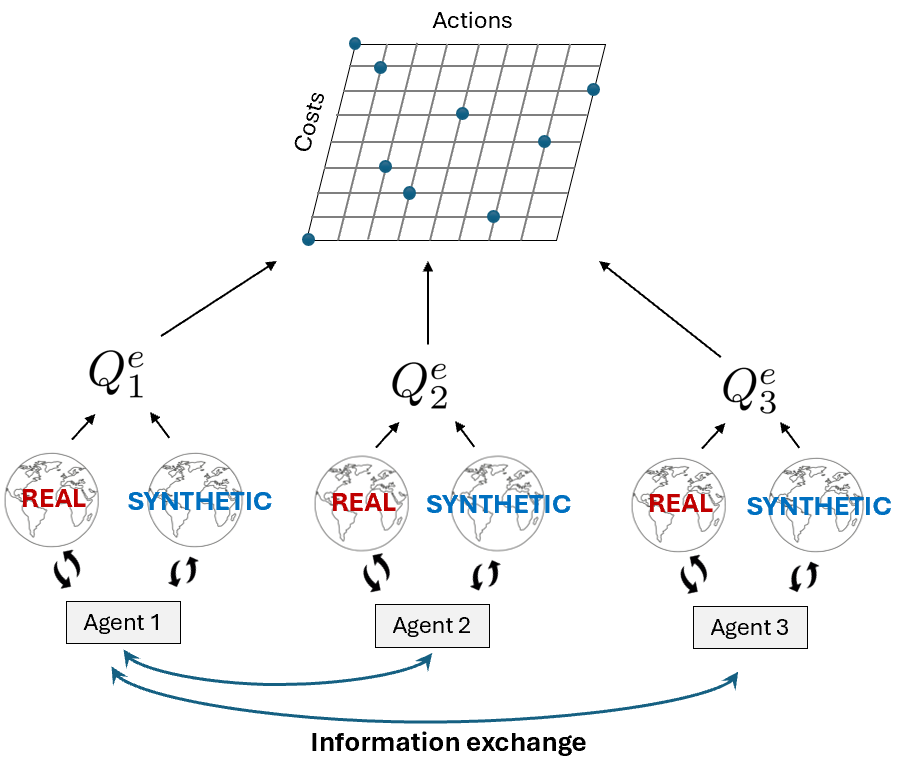}}}
    \caption{Comparison of different Q-learning algorithms.}
    \label{Fig:different_QL_algorithms}
\end{figure*} 

\section{System Model and Tools}
\label{sec:system_model}

\subsection{Markov Decision Processes}
\label{subsec:mdp}
A \textbf{single agent MDP} is characterized by a 4-tuple $\{\mathcal{S}_i, \mathcal{A}_i, p_i, c_i\}$, where $\mathcal{S}_i$ and $\mathcal{A}_i$ denote the state and action spaces of agent $i$. The state and action of agent $i$ at time $t$ are $s_{i,t}$ and $a_{i,t}$, respectively. The transition probability from $s_{i,t}$ to $s_{i,t}'$ under action $a_{i,t}$ is $p_{a_{i,t}}(s_{i,t}, s_{i,t}')$, which is stored in the probability transition tensor (PTT) $\mathbf{P}_i$, and the incurred cost is $c_i(s_{i,t}, a_{i,t})$. The agent $i$ aims to solve the following infinite-horizon discounted cost minimization equation (\textit{Bellman's optimality}) \cite{q_learning_survey}:
\begin{align}\label{Equ:single_agent_minimization}
    v_i^{*}(s_i) = \min_{\pi_i} v_{\pi_i}(s_i)= \min_{\bm{\pi}_i} \mathbb{E}_{\bm{\pi}_i} \sum_{t=0}^{\infty} \gamma^t c_i(s_{i,t}, a_{i,t}),
\end{align} 
where $v_{\pi_i}$ is the value function (VF) under policy $\pi_i$, $v_i^{*}$ is the optimal VF of agent $i$, and $\gamma \in (0,1)$ is the discount factor.

A \textbf{multi-agent MDP} generalizes a single-agent MDP to $N$ agents ($i$ = $1,2,...,N$), each with its own states and actions. The joint state at time $t$ is $\bar{s}_t = (s_{1,t}, \ldots, s_{N,t})$ and the joint action is $\bar{a}_t = (a_{1,t}, \ldots, a_{N,t})$. The joint state and action spaces are obtained as $\mathcal{S} = \bigtimes_{i=1}^N \mathcal{S}_i$ and $\mathcal{A} = \bigtimes_{i=1}^N \mathcal{A}_i$, where $\bigtimes$ denotes the Cartesian product. The transition from joint state $\bar{s}_t$ to $\bar{s}'_t$ under joint action $\bar{a}_t$ occurs with probability $p(\bar{s}_t, \bar{a}_t, \bar{s}'_t)$, which is stored in the joint probability transition tensor $\mathbf{\bar{P}}$, and incurs joint cost $c_i(\bar{s}_t, \bar{a}_t)$ for agent $i$. The optimization problem of the multi (joint) agent is as follows:
\begin{align}\label{Equ:multi_agent_minimization}
    \bar{v}^{*}(\bar{s}) = \min_{\bar{\pi}} \bar{v}_{\pi}(\bar{s})= \min_{\bar{\pi}} \mathbb{E}_{\bar{\pi}} \sum_{t=0}^{\infty} \gamma^t \sum_{i=1}^N c_i(\bar{s}_{t}, \bar{a}_{t}),
\end{align}
where $\bar{v}_{\bar{\pi}}$ is the joint value function under the joint policy $\bar{\pi}$ and $\bar{v}^{*}$ is the optimal joint value function. If agents are independent (i.e. no interactions), the joint PTT \( \mathbf{\bar{P}} \) is given by $\mathbf{\bar{P}}$ = $\bigtimes_{i=1}^N \mathbf{P}_i$; otherwise, \( \mathbf{\bar{P}} \) accounts for agent interactions.

\subsection{$Q$-Learning}
When the system dynamics (transition probabilities and cost functions) are unknown or non-observable, $Q$-learning can be used to solve (\ref{Equ:single_agent_minimization}) and (\ref{Equ:multi_agent_minimization}). \textbf{Single-agent $Q$-learning} solves (\ref{Equ:single_agent_minimization}) by learning the $Q$-functions of agent $i$, \(Q_i(s_i, a_i)\), using the following update rule \cite{q_learning_survey}:
\begin{equation}\label{Equ: Q-learning-update-rule}
    Q_i(s_i, a_i) \leftarrow (1 - \alpha) Q_i(s_i, a_i) + \alpha (c_i(s_i, a_i) + \gamma \min_{a_i' \in \mathcal{A}_i} Q_i(s_i', a_i')),
\end{equation}
where $\alpha \in (0,1)$ is the learning rate. Different exploration strategies, such as epsilon-greedy, softmax-exploration, or upper-confidence bound action selection, can be used to handle the exploration-exploitation trade-off to ensure sufficient sampling of different state-action pairs \cite{q_learning_survey}. The agent interacts with the environment and collects samples $\{s_i,a_i,s'_i,c_i\}$ to update $Q$-functions using (\ref{Equ: Q-learning-update-rule}). $Q$-functions converge to their optimal values with probability one, given certain conditions \cite{q_learning_convergence}. The optimal policy is $\pi_i^*(s_i) = \argmin_{a_i \in \mathcal{A}_i} Q_i^*(s_i, a_i)$, and the relationship between the optimal value function and optimal $Q$-functions is $v_i^*(s_i) = \min_{a_i \in \mathcal{A}_i} Q_i^*(s_i, a_i)$. $Q$-learning is an off-policy algorithm because it optimizes a target policy (the optimal policy being learned) while using a different policy, such as epsilon-greedy, for exploration and interaction with the environment. In addition, $Q$-learning effectively works with small to moderately large discrete state-action spaces.

Different MARL training strategies can be employed to extend single-agent $Q$-learning to multiple agents to solve (\ref{Equ:multi_agent_minimization}) depending on the availability of global information, observability of the joint state, or agents' ability to interact/communicate. The optimal joint policy and value functions can be obtained as $\bar{\pi}^*(\bar{s}) = \argmin_{\bar{a} \in \mathcal{A}} \bar{Q}^*(\bar{s},\bar{a})$, and $\bar{v}^*(\bar{s}) = \min_{\bar{a} \in \mathcal{A}} \bar{Q}^*(\bar{s},\bar{a})$ where $\bar{Q}^*$ is the optimal joint $Q$-functions. The sizes of the $Q$-tables for $Q_i$ and $\bar{Q}$ are $|\mathcal{S}_i| \times |\mathcal{A}_i|$ and $|\mathcal{S}| \times |\mathcal{A}|$, respectively. In this work, all agents are assumed to share the same state-action space size (i.e., $|\mathcal{S}_i| \times |\mathcal{A}_i|$ is the same for all $i$).

\subsection{Single-agent Multi-environment Mixed $Q$-Learning}\label{Subsec:memq}

MEMQ algorithm leverages the novel concept of \textit{digital cousins} proposed in \cite{pn_journal, ln_journal}. Digital cousins are multiple synthetically created, structurally related but distinct environments. Let $i$ subscript denote agent $i$. In MEMQ, the agent $i$ interacts with $K_i$ environments simultaneously: one real environment (Env$_1$), and $K_i-1$ synthetic environments (Env$_n$), where each Env$_n$ ($n > 1$) represents a unique digital cousin of the real environment. For example, the agent interacts with two environments in Fig.\ref{Fig:different_QL_algorithms}b: one real ('REAL') and one synthetic ('SYNTHETIC'), where the order $n$ is generic and not shown).

We herein briefly review the work of \cite{pn_journal}. The agent $i$ continuously estimates the PTT of the real environment in real-time via sample averaging from empirical state transitions collected under the $\varepsilon$-greedy exploration policy \cite{barto_sutton_rl}. The collected transitions are stored in the replay buffer and used to iteratively update the estimated PTT $\hat{\mathbf{P}}_i$ throughout training. We employ an $\varepsilon$-greedy policy instead of a purely random policy since the collected transition samples should reflect both exploration and the current learned policy. This approach produces Maximum Likelihood Estimate (MLE) for the PTTs unless additional structure exists \cite{barto_sutton_rl, colink_journal}. Let $\mathbf{\hat{P}}_i$ denote the estimated PTT. This estimate $\hat{\mathbf{P}}_i$ is then used to construct synthetic transition kernels for the digital cousins. Specifically, the $n$-step transition kernel of the $n^{\text{th}}$-order digital cousin is computed as $\hat{\mathbf{P}}_i^n$.\footnote{There are alternative ways to construct the PTT of synthetic environments such as employing graph symmetrization methods \cite{ln_journal}} Since $\hat{\mathbf{P}}_i$ is continuously updated during learning, the synthetic transition kernels are also continuously refined throughout training. The agent runs $K_i$ independent $Q$-learning algorithms across these environments, generating $Q$-functions $Q^{(n)}_i$ for each $\text{Env}_n$. These $Q$-functions are combined into an ensemble $Q$-function, $Q^e_i$, using the update rule:
\begin{equation}
    Q^e_i(s_i,a_i) = uQ^e_i(s_i,a_i)+(1-u)\sum_{n=1}^{K_i}w_i^{(n)}Q_i^{(n)}(s_i,a_i),\label{Equ:memq_update}
\end{equation}
where the weights of environments $w_i^{(n)}$ is determined using some adaptive weighting mechanism, and $u$ is the update ratio parameter \cite{pn_journal}. It has been shown that $Q^e_i$ converges to the optimal $Q$-functions in mean-square (i.e. \( Q^e_i \xrightarrow{m.s.} Q^* \) for all $i$) faster than the individual $Q$-functions $Q^{(n)}_i$ given a particular structure on update ratio $u$ \cite{pn_journal}. We emphasize that neither the synthetic $Q$-functions $Q_i^{(n)}$ nor their weighted combinations $\sum_{n=1}^{K_i} w_i^{(n)} Q_i^{(n)}$ generally converge to the optimal $Q$-function due to different underlying PTTs. Instead, the ensemble $Q$-function $Q_i^e$, which is obtained iteratively via update rule~(\ref{Equ:memq_update}), \textbf{does} converge to the optimal $Q$-function \cite{pn_journal}.

Herein, $K_i$ and $n$ are hyperparameters. While $K_i$ is chosen proportional to the state-action space size, selecting $n$ is more challenging. While the parameter $n$ can range from $1$ to $K_i$ as in (\ref{Equ:memq_update}), in practice, only a subset of these values is selected, since not all $K$ environments contribute useful samples. For $K_i=4$, there are many possible sets of orders (1,2,3,4 vs 1,2,3,5 vs 2,3,6,7 or similar). The goal is to use the fewest synthetic environments that provide the most diverse multi-time-scale information. To this end, we employ the coverage-coefficient environment selection algorithm that we developed for the single-agent case \cite{talha_spawc, talha_coverage_journal}, which selects the synthetic environments based on their sample coverage of the real environment. We also underline that, regardless of $K_i$, samples from synthetic environments cannot contain more information than real samples; hence, the real environment is \textbf{always} included in the MEMQ algorithm.

We use the notations $Q$-function and $Q$-table interchangeably. For instance, \( Q^{(n)}_i \) represents a generic \( Q \)-function for $\text{Env}_n$ as well as the \( Q \)-table that stores \( Q \)-functions for $\text{Env}_n$. Meanwhile, \( Q^{(n)}_i(s_i,a_i) \) refers to the \( Q \)-function for a specific state-action pair. 

\section{Decentralized Wireless Network with Multiple Networked Agents}

While our approach is broadly applicable to domains such as wireless network optimization, multi-agent games, and random network graph optimization (as demonstrated in the single-agent case \cite{pn_journal, ln_journal}), in this work we concentrate on decentralized wireless networks. This setting provides a representative example of real-world network complexities and challenges, while the framework remains readily adaptable to games, network graphs, and other applications.

\begin{figure}[t]
    \centering
    \includegraphics[width=0.45\textwidth]{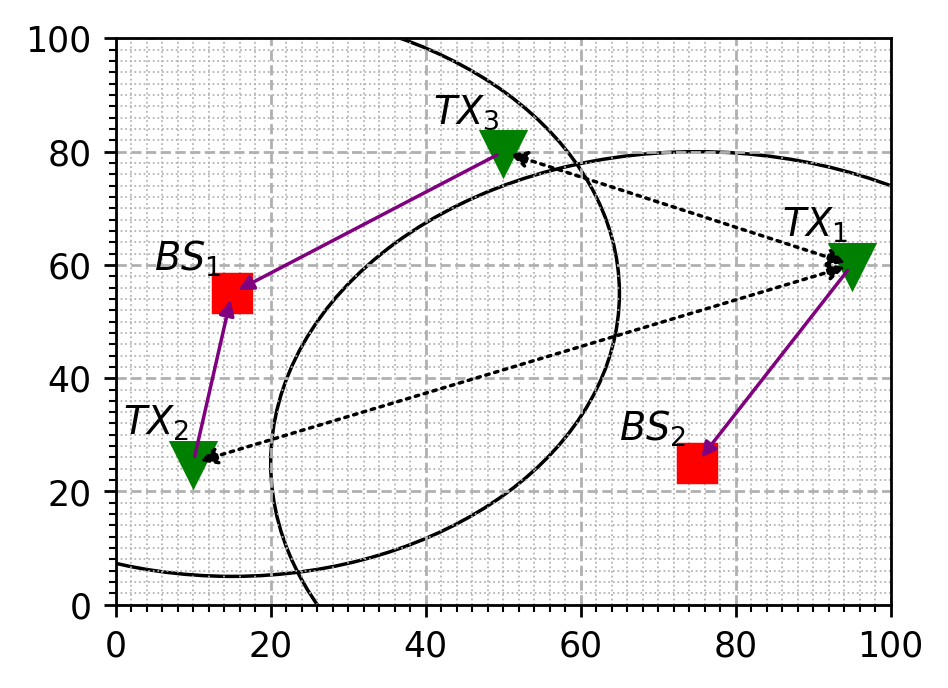}
    \caption{Example wireless network setting}
    \label{Fig:eaxmple_wireless_network}
\end{figure}

We consider a grid environment of size $L \times L$ with cells of size $\Delta_{L} \times \Delta_{L}$. There are $N_T$ mobile transmitters (TXs) and $N_B$ stationary base stations (BSs), where each TX can move randomly in four directions (U, R, D, L) or stay stationary, all with equal probability. TXs always remain on grid points. Each BS$_j$ has a circular coverage area with radius $r_j$; a TX can connect to BS only on or within this area. Fig.\ref{Fig:eaxmple_wireless_network} shows an example wireless network with 3 TXs and 2 BSs, where red squares are BSs, green triangles are TXs, black solid circles show coverage areas, purple solid arrows represent a particular TX-BS association, black dotted arrows represent the communications between TXs, and TX$_1$ is assumed to be the leader TX that controls the coordination among agents (i.e. TXs communicate with \textbf{only} TX$_1$ when necessary). Herein, the leader TX is defined as the transmitter with the largest coverage area (radius $r_j$), as it can communicate with and broadcast to the largest portion of the network, ensuring better connectivity to base stations and other TXs. We underline that many hierarchical sensor and wireless networks have leader transmitters or nodes \cite{leader_node_1}. We emphasize that although leader selection introduces minor overhead, the leader need not be fixed and can be reassigned across simulations; as long as a central agent with sufficiently large coverage exists, our analysis remains valid. Each TX can continuously measure (sense) the aggregated received signal strength (ARSS) due to all other TXs. This is practical, as most wireless devices can easily obtain ARSS measurements, unlike CSI information, which requires specialized hardware \cite{molisch2012wireless}. We now describe the RL framework for the given wireless network.

\subsubsection{\textbf{Agent}} TXs are the decision-making agents; hence, there are $N_T$ agents. While TXs operate in a decentralized manner as the joint state information is not available (see next section), they are networked and can coordinate through the leader TX when coordination is necessary (see Section IV).

\subsubsection{\textbf{State}} The individual state of TX$_i$ is a triplet $s_i = (x_i, y_i, I_i)$, where $(x_i, y_i)$ is its location with $x_i, y_i \in [0, \Delta_L, 2\Delta_L, ..., L]$, and $I_i$ is the ARSS. Since ARSS may be continuous, we quantize \(I_i\) to the nearest value in \([I_{\text{min}}, I_{\text{max}}]\), which is discretized into equal steps of size \(\Delta_I\). The transition from $(x_i, y_i)$ to $(x_i', y_i')$ depends on the random movement of TX$_i$, while the transition from $I_i$ to $I_i'$ is influenced by several factors such as the random movement of all TXs, the distances between TXs, $L$, and $I_{thr}$. The joint state $\bar{s}$ is the concatenation of individual states as $\bar{s} = (s_1, s_2, ..., s_{N_T})$, and is \textbf{not} available to the agents (i.e. agents do not have access to each others' states).

The probability of transitioning from state $s_i$ to $s_i'$ is:
\begin{equation}
    P(s_i' \mid s_i) 
    = P_x(x_i' \mid x_i) \; P_y(y_i' \mid y_i) \; P_I(I_i' \mid I_i),
    \label{Equ: state_transition_probs}
\end{equation}
where $P_x(\cdot)$ and $P_y(\cdot)$ denote the probabilities of the user’s $x$- and $y$-coordinates performing a random walk over the possible directions. The term $P_I(\cdot)$ is the probability of transitioning from the ARSS value $I_i$ to $I_i'$, which is influenced by the location of the users and their random movements, their local ARSS measurements, and the discretization and quantization of the continuous ARSS into the discrete representation.

The state-space is divided into two parts: \textit{coordinated} and \textit{uncoordinated} states. The joint state $\bar{s}$ is \textit{coordinated} if there is at least one strong ARSS measurement in the wireless network, i.e., there is at least one agent $i$ such that $I_{i} > I_{thr}$, where $I_{thr}$ is the ARSS threshold; otherwise, it is \textit{uncoordinated}. Thus, if at least one TX identifies the state as coordinated, it informs TX$_1$, which then broadcasts this information to all TXs. Importantly, coordination is triggered by the wireless environment rather than imposed by the algorithm. TXs operate independently in uncoordinated states and coordinate only when strong ARSS measurements indicate significant inter-agent interactions. Herein, high ARSS indicates an agent receives strong signals from nearby users, as signal strength decays with distance (path loss). Conversely, low ARSS occurs when the agent is isolated or far from other users, and receives weak signals. In real wireless networks, ARSS is analogous to perceived voice quality, which can degrade in dense networks due to interference and latency. In such high-density situations, corresponding to \textit{coordinated states}, users can benefit from coordination via the central controller (i.e. the leader agent) by reducing interference, collisions, and latency.

We note that there are alternative approaches to determining coordinated states, such as based on predefined states \cite{coordinated_State_5} or proximity between agents \cite{coordinated_State_8}. Our approach is similar to \cite{coordinated_State_4}, which also uses local RSS measurements to determine coordinated states. However, in our work, coordination is based on sufficiently high ARSS values. In addition, coordination is possible even if not all agents measure high ARSS levels. This makes our scheme robust against ARSS measurement inaccuracies in some agents, and scalable by reducing dependency on each agent's measurement.

We denote $|\mathcal{S}_C|$ and $|\mathcal{S}_U|$ as the number of coordinated and uncoordinated states such that $|\mathcal{S_C}|+|\mathcal{S_U}|=|\mathcal{S}|$. We theoretically derive the optimal value of $I_{thr}$ that minimizes the lower bound on the probability of misdetection (i.e., detecting the state as coordinated while it is actually uncoordinated or vice versa) in the next section.

\setlength{\textfloatsep}{5pt}

\subsubsection{\textbf{Action}}
The action set of TX$_i$ is defined as 
\[
\mathcal{A}_i = \{(m,b) \mid m \in \{0,1\}, \; b \in \{1,2,\ldots,N_B\}\},
\] 
where $m=0$ denotes staying stationary and $m=1$ denotes initiating movement. If $m=1$, the next position is sampled from a random walk. The index $b$ specifies the BS to which TX$_i$ connects; actions corresponding to BSs outside coverage are invalid and incur infinite cost. The joint action is $\bar{a} = (a_1, a_2, \ldots, a_{N_T})$ for each $a_i \in \mathcal{A}_i$. Herein, the action of each TX$_i$ directly affects its state transitions in (\ref{Equ: state_transition_probs}).

\subsubsection{\textbf{Cost function}} The individual cost function of TX$_i$ is as follows: 
\begin{align}
    c_i = \beta_1\frac{P_{i}}{\log_2(1 + \operatorname{SNR}_i)} + \beta_2\mathcal{Q}(\sqrt{\text{SNR}_i}) + \beta_3\mathcal{Q}(\frac{I_{thr}-I_{i}}{\sigma}),\label{Equ: individual_cost_func}
\end{align}
where $\operatorname{SNR}_i = \frac{P_{i}}{d_i^2(n + I_i)}$, $d_i$ is distance between TX$_i$ and the BS it is connected to, $P_{i}$ is the transmission power of TX$_i$ which is fixed but may differ among agents, and $n$ is the zero-mean Gaussian noise as $n \sim \mathcal{N}(0,\sigma^2)$. Herein, the first component is the \textit{efficiency cost}. A higher SNR with constant transmitting power improves the effective data rate and reduces the efficiency cost. The second component is the \textit{reliability cost}. The $\mathcal{Q}(x) = \frac{1}{2}\operatorname{erfc}\left( \frac{x}{\sqrt{2}} \right)$ represents the inverse relationship between SNR and bit error rate. Thus, higher SNR reduces the bit error rate and improves transmission reliability. The third component is the \textit{fairness cost}. If the measured local ARSS exceeds the threshold $I_{thr}$, it suggests that some TXs are close to each other and may compete for the same resources (time slots, frequencies), potentially overloading some BSs while underutilizing others, which increases the fairness cost. The constants $\beta_1, \beta_2, \beta_3$ are the cost weights such that $\beta_1 + \beta_2 + \beta_3 = 1$. The joint cost is the sum of individual costs: $\bar{c} = \sum_{i=1}^{N_T} c_i(\bar{s}_{t}, \bar{a}_{t})$. We emphasize that the individual cost function of each TX in (\ref{Equ: individual_cost_func}) relies solely on local ARSS observations and does not depend on the knowledge of other TXs. 

This cost function has several key properties. It is non-negative and bounded. It is highly non-linear and non-monotonic in some parameters (e.g., \( P_i \)), but monotonic in others (e.g., SNR and \( I_{thr} \)). {Its non-convexity (due to the Gaussian $\mathcal{Q}$-function)} presents optimization challenges. The effect of different costs and parameters can be controlled through the weights \( \beta_i \).

\section{Algorithm}
\label{sec:algorithm}

\subsection{Preliminaries and initialization}

We describe our algorithm (Algorithm 1) using a three-agent wireless network ($N_T=3$) to explain key details. However, our approach works for an arbitrary number of agents, as the wireless network setting in the previous section is described for arbitrary $N_T$. Our simulations will explicitly consider larger values of $N_T$. Fig.\ref{Fig:different_QL_algorithms} illustrates the high-level comparison between standard $Q$-learning \cite{q_learning_survey}, single-agent MEMQ \cite{pn_journal}, and the proposed multi-agent MEMQ, where blue arrows show the direction of information exchange between agents and leader agent.

\begin{algorithm}[t]
\small
\caption{Multi-Agent MEMQ (M-MEMQ)}
\hspace*{\algorithmicindent} \textbf{Inputs:} $l$, $\gamma$, $\alpha_t$, $\zeta_t$, $u_t$, $\mathbf{b}_{i,0}$, $Q_i^{(n)}$, $Q^{e}_i$, wireless parameters \\
\hspace*{\algorithmicindent} \textbf{Outputs:} $\bar{Q}, \bar{\pi}$
\begin{algorithmic}[1]
    \For{$t = 1$ to $T$}
        \For{each TX$_i$ \textbf{simultaneously}}
            \State Reinitialize $\mathbf{b}_{i,t}$ if $\operatorname{mod}(t,l) = 0$
            \State Interact with Env$_1$, collect samples \(\{s_i, a_i, s_i', c_i\}\), and store in buffer $\mathcal{B}_i$
            \State Sample \(\{s_i, a_i, s_i', c_i\}\) randomly from $\mathcal{B}_i$ to update $\mathbf{P}_i$ and the PTT of synthetic environment  
            \State Determine if current state is coordinated based on local ARSS, and if so, estimate it using the Bayesian approach
            \State Update the local and joint $Q$-functions as in Table \ref{Table:q_function_update_4_cases}
        \EndFor
    \EndFor
\State $\bar{\pi}(\bar{s}) \gets \operatorname*{argmin}_{\bar{a}'}\bar{Q}(\bar{s},\bar{a}')$
\end{algorithmic}
\label{Algorithm: multi_agent_memq}
\end{algorithm}

The inputs to Algorithm 1 are \textit{algorithm parameters} (trajectory length $l$, discount factor $\gamma$, learning rate $\alpha_t$, exploration probability $\zeta_t$, update ratio $u_t$, initial belief vectors $\mathbf{b}_{i,0}$ for agent $i$ at $t=0$ (defined in Section IV-E), number of iterations $T$) and \textit{wireless network parameters} ($L$, $\Delta_L$, $N_T$, $N_B$, $r_j$, $I_{min}$, $I_{max}$, $\Delta_I$, $\beta_1, \beta_2, \beta_3$, $P_i$, $k$, $I_{thr}$, $\sigma$). The outputs are the joint $Q$-table, $\bar{Q}$, and the corresponding joint policy, $\bar{\pi}$. 

Each TX interacts with 2 environments as shown in Fig.\ref{Fig:different_QL_algorithms}: the real environment (Env$_1$) and a single synthetic environment. The order of the synthetic environments for three TXs is chosen to be 2, 3, and 5, respectively, which are determined using the coverage-coefficient environment selection algorithm as explained in Sec.\ref{Subsec:memq}. Hence, each TX keeps three different \textbf{local} $Q$-functions (i.e. $Q$-tables) as shown in Table \ref{tab:local_Q_tables}: two individual and one ensemble $Q$-functions. The leader TX, TX$_1$, also stores the joint $Q$-table $\bar{Q}$, which is updated based on the local $Q$-functions of all agents. We emphasize that $K=2$ (i.e. one real and one synthetic environment for each agent) yields optimal performance for the given wireless network, yet our approach is compatible with any $K$; other wireless networks such as in \cite{colink_journal, ln_journal} may require a larger $K$ (and consequently different synthetic environments). 

At the beginning of the algorithm, transmitters and base stations are also located randomly within the grid, while ensuring no collisions and sufficient distance between BSs. 

\subsection{Creating synthetic environments} Each TX$_i$ independently interacts with the real environment Env$_1$, collects online samples \(\{s_i, a_i, s_i', c_i\}\) to update its Q-functions for the real environment, and stores this data in their buffer $\mathcal{B}_i$. Then, each TX$_i$ randomly samples offline mini-batches from $\mathcal{B}_i$ to update their individual estimated PTT for Env$_1$ ($\mathbf{\hat{P}}_i$), and simultaneously builds and updates the PTT of its synthetic environment ($\mathbf{\hat{P}}_i^n$) by taking the matrix power of $\mathbf{\hat{P}}_i$ as explained in Sec.\ref{Subsec:memq}. In this way, $Q$-functions are updated using online data, while offline samples from the buffer are used exclusively for kernel (PTT) estimation and updates. We emphasize that creating synthetic environments is not a one-time process; the PTTs of synthetic environments are \textbf{continuously} updated as each TX continues sampling from the real environment. Thus, the accuracy of the estimated synthetic environments improves over time.

We also reset belief updates and reinitialize belief vectors every $l$ step to mitigate error accumulation from potentially erroneous initial belief vectors, noisy $Q$-functions, incorrect $Q$-function updates, or numerical errors during matrix power operations and improve adaptation to new observations. This section of the algorithm is given between lines 3 and 5.

\renewcommand{\arraystretch}{1.4}
\begin{table}[t]
    \centering
    \begin{tabular}{|c|c|c|c|}
        \hline
        & \multicolumn{2}{c|}{Local $Q$-tables} & \multicolumn{1}{c|}{Joint $Q$-table} \\
        \cline{2-3}
        & Individual & Ensemble & \\
        \hline
        TX$_1$ & $Q_1^{(1)}$, $Q_1^{(2)}$ & $Q_1^{e}$ & $\bar{Q}$ \\
        \hline
        TX$_2$ & $Q_2^{(1)}$, $Q_2^{(3)}$ & $Q_2^{e}$ & - \\
        \hline
        TX$_3$ & $Q_3^{(1)}$, $Q_3^{(5)}$ & $Q_3^{e}$ & - \\
        \hline
    \end{tabular}
    \caption{Local and joint $Q$-tables in Algorithm \ref{Algorithm: multi_agent_memq}}
    \label{tab:local_Q_tables}
\end{table}

\renewcommand{\arraystretch}{1.7}
\begin{table*}[ht]
\footnotesize
\centering
\begin{tabular}{|c|c|c|}
\hline
\centering\textbf{State transition} & \textbf{Update rules}\\
\hline
U $\rightarrow$ U &
$Q_{i}^{(n)}(s_i,a_i) \leftarrow \text{Q-learning}, \quad Q^{e}_i(s_i,a_i) \leftarrow$ MEMQ, \quad $\bar{Q}(\bar{s},\bar{a}) = \sum_{i=1}^{N_T} Q_{i}^e(s_i,a_i)$.\\
\hline
U $\rightarrow$ C &
$Q_{i}^{(n)}(s_i,a_i) \leftarrow (1-\alpha)Q_{i}^{(n)}(s_i,a_i) + \alpha\big[c_i(s_i,a_i) +  \frac{\gamma}{3} \min_{\bar{a}'} \bar{Q}(\bar{s}', \bar{a}')\big]$, \quad $Q^{e}_i(s_i,a_i) \leftarrow$ MEMQ\\
\hline
C $\rightarrow$ U &
$\bar{Q}(\bar{s},\bar{a}) \leftarrow (1-\alpha)\bar{Q}(\bar{s},\bar{a}) + \alpha \sum_{i=1}^3\big[c_i(\bar{s},\bar{a}) + \gamma \min_{a_i'} Q_{i}^{e}(s_i', a_i')\big]$\\
\hline
C $\rightarrow$ C &
$\bar{Q}(\bar{s},\bar{a}) \leftarrow (1-\alpha)\bar{Q}(\bar{s},\bar{a}) + \alpha\big[\sum_{i=1}^3c_i(\bar{s},\bar{a}) + \gamma \min_{\bar{a}'} \bar{Q}(\bar{s}', \bar{a}')\big]$\\
\hline
\end{tabular}
\caption{Q-function update rules for different cases (time index subscript $t$ is dropped for simplicity)}
\label{Table:q_function_update_4_cases}
\end{table*}

\subsection{Joint state estimation in Coordinated States}

While TXs can act independently in uncoordinated states, they need to coordinate in coordinated states to minimize the joint cost and update the joint $Q$-table. However, each TX has access to only local knowledge -- they can continuously measure ARSS due to other TXs without further access to their individual states, actions, or costs. To this end, agents can estimate the joint state based on their local ARSS observations, and share limited information to obtain the joint action and costs. This approach, while not always optimal, ensures some level of privacy and significantly reduces information exchange cost \cite{based_on_communication_strategies}. 

In particular, agents use a Bayesian approach, akin to standard MAP estimation, to estimate the joint state in coordinated states based on their local ARSS observations. For TX$_1$, this involves estimating the joint state of TX$_2$ and TX$_3$ using the local ARSS measurement $I_{1,t}$ at time $t$. Let \( (s_{2}, s_{3}) \) be the actual states of TX\(_2\) and TX\(_3\) at time \( t \), and \( (\hat{s}_{2}, \hat{s}_{3}) \) their estimates by TX\(_1\). Then,
\begin{align}
    \hat{s}_{2}, \hat{s}_{3} &= \underset{s_{2}, s_{3}}{\operatorname{argmax}} \hspace{4pt} p(s_{2}, s_{3} \mid I_{1,t}), \label{Equ: argmax_0}\\
    &= \underset{s_{2}, s_{3}}{\operatorname{argmax}} \hspace{4pt} p(I_{1,t} \mid s_{2}, s_{3}) \mathbf{b}_{1,t}(s_{2}, s_{3}),\label{Equ: argmax}
\end{align}
where the conditional distribution $ p(I_{1,t} \mid s_{2}, s_{3})$ follows a multivariate Normal distribution with mean vector \( \bm{\mu}_{1,t} \) and covariance matrix \( \bm{\Sigma}_{1,t} \), and \( \mathbf{b}_{1,t} \) denotes TX\(_1\)'s belief vector at time \( t \) -- the probability that TX\(_1\) believes TX\(_2\) and TX\(_3\) are in states $s_{2}$ and $s_{3}$, respectively. Herein, the parameters \( \bm{\mu}_{1,t} \) and \( \bm{\Sigma}_{1,t} \) depend on the ARSS measurements up to time \( t \). The computation of these parameters are given in the supplementary material.

Then, the estimated joint state of the system by TX$_1$ is given by: $(s_{1}, \hat{s}_{2}, \hat{s}_{3})$. Furthermore, the belief probabilities for the states \( (s_{2}, s_{3}) \) is updated as:
\begin{align}
    \mathbf{b}_{1,t+1}(s_{2}, s_{3}) &\leftarrow p(I_{1,t} \mid s_{2}, s_{3}) \mathbf{b}_{1,t}(s_{2}, s_{3}), \label{Equ: argmax_belief} \\
    \mathbf{b}_{1,t+1}(s_{2}, s_{3}) &\leftarrow \text{softmax}\left(\mathbf{b}_{1,t+1}(s_{2}, s_{3})\right). \label{Equ: softmax}
\end{align}

The joint state estimation for other TXs is done in a similar way; hence, each TX$_i$ at time $t$ has its own joint state estimate and a corresponding belief vector. The accuracy of joint state estimates varies among TXs depending on the quality of ARSS measurements, mean and covariance parameters, and initial belief vectors. This section of the algorithm is given in line 7.

\begin{remark} \normalfont
We emphasize that the argmax operation in (\ref{Equ: argmax}) is not an exhaustive discrete search over the entire state-space. The search at time $t$ is restricted to a small subset of the state-space \(\{s_{2,t}: \|s_{2,0} - s_{2,t}\|_2 \leq \Delta\} \times \{s_{3,t}: \|s_{3,0} - s_{3,t}\|_2 \leq \Delta\)\}, where $\Delta = \delta_0+2\Delta_L\operatorname{min}(l, \operatorname{mod}(t,l))$ and the distance between the first estimated and true state is at most $\delta_0$ \cite{talha_icassp_25}. Thus, the search space size is independent of the joint state-space size. We will numerically show that estimated states converge to the true states for modest values of $\Delta$. A small \( \delta_0 \) can be determined by analyzing the historical behavior of TXs, or leveraging domain knowledge. Without such knowledge, smaller trajectory lengths \( l \) or smaller grid size \( \Delta_L \) also helps.
\end{remark}

\begin{remark} \normalfont
    If the belief vector in (\ref{Equ: argmax_belief}) has also a multivariate Gaussian form as $\mathbf{b}_{1,t} \sim N(\bm{\mu}'_{1,t}, \bm{\Sigma}'_{1,t})$, then the uncertainty in the updated belief vector $\mathbf{b}_{1,t+1}$ monotonically decreases over time, i.e., $\bm{\Sigma}'_{1,t+1} \preceq \bm{\Sigma}'_{1,t}$. This result can be shown using the properties of multivariate Gaussian. On the other hand, the updated mean depends on the local ARSS measurements, which vary over time as TXs move and channel conditions change, and hence may not change monotonically over time.
\end{remark}

\subsection{Local and Joint Q-function updates}

In Algorithm~\ref{Algorithm: multi_agent_memq}, $Q$-function updates depend on transitions between coordinated (\textbf{C}) and uncoordinated (\textbf{U}) states. Hence, they are 4 different cases as shown in Table~\ref{Table:q_function_update_4_cases}. 

In \textbf{{U~$\to$~U}} step, each agent independently updates its local $Q$-functions ($Q_i^{(n)}, Q_i^e$). Agents update their local $Q$-functions across environments ($Q_i^{(n)}$) using standard $Q$-learning, and independently update their ensemble $Q$-functions ($Q_i^e$) via the single-agent MEMQ algorithm. To this end, several alternatives exist \cite{talha_eusipco, pn_journal, ln_journal}. While these algorithms share similar goals, they differ in assumptions and network constraints. We adopt \cite{pn_journal} as the single-agent MEMQ algorithm due to its broad applicability with minimal assumptions. For networks with specific constraints, one may instead use \cite{talha_eusipco, ln_journal}. We also assume that the joint $Q$-function satisfies an additivity property over the agents’ ensemble $Q$-functions as no information exchange takes place. Our formulation differs from prior work \cite{yu2014coordinated, sum_of_Q_functions, VDN} in that we aggregate local \emph{ensemble} $Q$-functions ($Q_i^e$) rather than traditional individual $Q$-functions ($Q_i^{(n)}$). To establish convergence of this update rule, one can directly apply the single-agent MEMQ convergence proof \cite{pn_journal, ln_journal} to each agent independently.

In \textbf{{C~$\to$~C}} step, the leader sums the agents reported local costs and uses this total to update $\bar{Q}$ from the previous iteration. In \textbf{{C~$\to$~U}} step, the leader agent transfers (backs up) each agent’s local ensemble $Q$-functions into $\bar{Q}$. As the system stays in coordination mode, individual agent-specific $Q$-functions are not updated. In \textbf{{U~$\to$~C}} step, the leader agent transfers $\bar{Q}$ into each agent’s $Q$-function, assigning an equal share of the expected discounted cost (e.g., $\tfrac{1}{3}$ for three agents). For these three steps, the system either operates centralized or with globally shared information. Thus, convergence follows standard $Q$-learning \cite{q_learning_convergence} by treating all agents as a single aggregated agent with global joint information.

When coordination is required, each TX first estimates the joint state from its local ARSS observations (Sec. IV-D) and shares the belief probability with the leader TX$_1$. The leader then selects the joint state with the highest probability and broadcasts the joint action that minimizes the joint $Q$-function. Each TX executes its action and reports costs and $Q$-values to the leader, which then updates the joint $Q$-function. This section of the algorithm is given in line 7.

Similar update rules to Table II have been used in prior MARL studies~\cite{kok2004sparse, yu2014coordinated, coordinated_State_4, coordinated_State_5}. Our algorithm extends them by updating both the local individual $(Q_i^{(n)})$ and local ensemble $Q$-functions $(Q_i^e)$.

\begin{remark} \normalfont
    The cost of information exchange over $T$ iterations among $N_T$ agents is proportional to $O(N_T\max\{T, |\mathcal{S}_i||\mathcal{A}_i|\})$, where $O$ is the Big O notation, and $i$ is a generic TX. \cite{talha_icassp_25}
\end{remark}

This result shows that the cost of information exchange scales linearly with the number of agents as the leader TX coordinates and simplifies the coordination among agents. Additionally, the cost depends only on the individual state-action space size, rather than the joint state-action space size. These two factors allow for accommodating more agents without a significant increase in communication costs.

We emphasize the partially decentralized nature of our algorithm and highlight key distinctions from fully centralized methods \cite{MARL_SURVEY_0, MARL_SURVEY_1, MARL_SURVEY_2, marl_challenges}. In our approach, agents do not have access to the global joint state, whereas fully centralized methods assume complete global state observability. Second, the leader agent is active and computes the joint actions only when coordination is triggered (based on local ARSS measurements); otherwise, agents act independently without communication or information exchange. Communication is therefore sparse and conditional in contrast to continuous, full-network communication in centralized methods. As a result, our communication cost scales linearly with the number of agents, while centralized schemes often incur exponential growth with the joint state–action space.

\section{Theoretical Analysis}

We provide several theoretical results for Algorithm~\ref{Algorithm: multi_agent_memq}, which are summarized in Table~\ref{Table:theory_summary}. All proofs are given in the supplementary material. While we exploit key assumptions to prove the results, these assumptions are not invoked in the numerical results (Section~\ref{sec:numerical_results}) which do serve to validate the trends suggested by the theory.

\begin{table*}[t]
\centering
\caption{Summary of stability conditions and corresponding theoretical guarantees.}
\label{Table:theory_summary}
\renewcommand{\arraystretch}{1.15}
\begin{tabular}{|p{4.0cm}|p{2.0cm}|p{4.8cm}|p{3.8cm}|}
\hline
\textbf{Condition} & \textbf{Result} & \textbf{Guarantee} & \textbf{Importance} \\ \hline

$\mathbb{E}[\mathcal{X}^{(n)}_{t,i}\mathcal{X}^{(m)}_{k,j}] = 0$, $n$$\neq$$ m$, $t\neq k$
&
Proposition~1
&
$\mathbb{V}\!\left(\bar{Q}_t-\sum_i Q_i^*\right)
\leq
\sum_i \frac{f_i(\lambda_i,t)}{K_i}$
&
Estimation variance reduces with number of environments $K_i$
\\ \hline

$\mathbb{E}[\mathcal{X}^{(n)}_{t,i}\mathcal{X}^{(m)}_{k,j}] = 0$, $n\neq m$
&
Corollary~1
&
$\lim\limits_{t\to\infty}
\mathbb{V}\!\left(\bar{Q}_t-\sum_i Q_i^*\right)
\leq
\frac{1-u}{1+u}\sum_i \lambda_i^2$
&
Larger update ratio $u$ improves asymptotic stability and reduces estimation variance.
\\ \hline

$\mathrm{Var}(\mathcal{X}^{(n)}_{t,i}) < \infty$
&
Corollary~\ref{Corollary_2}
&
$\bar{Q}_t(\bar{s},\bar{a})
\xrightarrow{\mathrm{m.s.}}
\sum_i Q_i^*(s_i,a_i)$
&
Joint $Q$-functions converge in mean square to optimal solution
\\ \hline

$|\epsilon_{i,t}^{(n)}(s_i,a_i)|
\leq
\theta_i^{(n)}(s_i,a_i)$
&
Corollary~3
&
$\displaystyle
t_\beta
=
\frac{1}{\log(u)} \log\!\left(
1-\frac{\beta}{
\sum_i \sum_n \theta_i^{(n)}
}
\right)
$
&
Provides convergence-rate
and sample-complexity guarantee
\\ \hline
\end{tabular}
\end{table*}
\subsection{Probabilistic Analysis}

We first provide several probabilistic results.

\begin{assumption} \normalfont
The following probabilistic results assume finite-variance $Q$-function estimation errors satisfying either the zero-mean assumption in (\ref{Equ: distribution_assumption}) or the generalized weighted-zero-mean assumption in (\ref{Equ: new_distribution_assumption}). Proposition~1 and Corollary~1 additionally assume independence across environments.
\end{assumption}

The probabilistic analysis is based on the following assumption on the $Q$-function estimation errors:
\begin{align}
\mathcal{X}^{(n)}_{t,i}(s_i,a_i)
=
Q_{t,i}^{(n)}(s_i,a_i)
-
Q_i^*(s_i,a_i)
\sim
D_{i,n}\!\left(0,\tfrac{\lambda_{i,n}^2}{3}\right),
\label{Equ: distribution_assumption}
\end{align}
where \(D_{i,n}\) is a zero-mean distribution with variance \(\lambda_{i,n}^2/3\).

This assumption is widely employed in RL literature: (i) it is used for non-MEMQ algorithms with $D_{i,n}$ being uniform or normal for $n=1$ case in \cite{uniform_assump_1, randomized_double_q}; (ii) it is generalized to $n>1$ with no assumptions on $D_{i,n}$ for MEMQ algorithms in \cite{talha_icassp_25, pn_journal, ln_journal}; and (iii) it is employed in MEMQ algorithms with $D_{i,n}$ being uniform in \cite{talha_coverage_journal}. These assumptions are also theoretically and numerically \textbf{justified} in \cite{talha_icassp_25, pn_journal, ln_journal}. This assumption is generalized to multiple agents in this work. We emphasize that this assumption is introduced solely to \textbf{facilitate} the theoretical analysis yet it can be relaxed as:
\begin{align}
\hspace{-7pt}\mathcal{X}^{(n)}_{t,i}(s_i,a_i)\!=\!Q_{t,i}^{(n)}(s_i,a_i)\!-\!Q^{*}_i(s_i,a_i)\!\sim\!D_{i,n} \left(\!\mu_{i,n},\! \tfrac{\lambda_{i,n}^2}{3}\!\right),\hspace{-4pt}
\label{Equ: new_distribution_assumption}
\end{align}
with the condition that
\begin{align}
\sum_{n=1}^K w_{i}^{(n)} \mu_{i,n} = 0,
\end{align}
where \( w_{i}^{(n)} \) denotes the weight of the \( n^{\text{th}} \) synthetic environment in the MEMQ algorithm of agent~\( i \). This assumption is less restrictive than (\ref{Equ: distribution_assumption}), as it only requires the weighted convex combination of biases to be zero, allowing individual means to be non-zero. Herein, the signs of $\mu_{i,n}$ depends on $n$. Thus, (i) over and underestimation effects cancel each other across different $n$, and (ii) higher-bias $Q$-functions receives smaller weights $w_i^{(n)}$ as shown in \cite{pn_journal}, ensuring the condition holds in practice. Our theoretical results \textbf{hold} under both cases. We also emphasize that both (\ref{Equ: distribution_assumption}) and (\ref{Equ: new_distribution_assumption}) assume that the $Q$-function estimation \emph{errors}, not the $Q$-function updates themselves follow a random distribution with finite variance.

We herein leverage probabilistic results in \cite{pn_journal}, which were derived for the single-agent MEMQ to derive our results for multi-agent MEMQ. These results hold for any number of agents ($N_T$) and number of environments ($K_i$).

\begin{proposition} \normalfont
    Let the estimation errors $\mathcal{X}^{(n)}_{t,i}$ be independent across different $n$ for all $i,t$. Let $\bar{s}$ be an uncoordinated joint state. Then, the following holds for all $t$:
    \begin{align}
        \mathbb{V}\Big(\bar{Q}_t(\bar{s},\bar{a}) - \sum_{i=1}^{N_T} Q_i^*(s_i,a_i)\Big) \leq \sum_{i=1}^{N_T} \frac{f_i(\lambda_i, t)}{K_i},\label{Equ:Prop1}
    \end{align}
    where $\mathbb{V}$ is the variance operator, $\lambda_i = \max\limits_{n}\lambda_{i,n}$ is the estimation error variance of the worst environment for agent $i$, and $f_i$ is some function of the parameter $\lambda_i$. 
\end{proposition}

This proposition has several implications. First, as the number of environments ($K_i$) increases, the upper bound on the estimation error variance decreases, which is consistent with the variance-reduction property of ensemble RL algorithms. Second, even if some agents employ a small number of environments, the estimation error variance can still be reduced by employing a larger number of environments for other agents, which differs from the single-agent case \cite{pn_journal}, where employing a smaller number of environments always lead to a larger estimation variance. Third, the function $f_i$ can be fine-tuned to further reduce the upper bound (see \cite{pn_journal}). Finally, this result gives us the behavior of the estimation error variance in the finite $t$ case. We emphasize that this bound characterizes the variance trend of the $Q$-function estimation error under the idealized assumption that $\mathcal{X}^{(n)}_{t,i}(s_i,a_i)$ are independent across different $n$ for each $t$ and $i$ \cite{pn_journal}. Increasing $K$ reduces the error but does not completely eliminate it in implementation, as there is a diminishing return as $K$ gets large, and the parameter $\lambda_i$ is a function of $ K$.

\begin{corollary} \normalfont
    The upper bound in (\ref{Equ:Prop1}) can also be expressed as a function of the update ratio $u$ asymptotically in time as:
    \begin{align}
        \lim_{t \rightarrow \infty} \mathbb{V}\Big(\bar{Q}_t(\bar{s},\bar{a}) - \sum_{i=1}^{N_T} Q_i^*(s_i,a_i)\Big) \leq  \frac{1-u}{1+u}\sum_{i=1}^{N_T} \lambda_i^2.
    \end{align}
\end{corollary}

This result indicates that even with a large number of agents ($N_T$), the estimation error variance can be reduced by choosing a large $u$ or using a time-varying $u_t$ (such that $u_t \xrightarrow[]{t \rightarrow \infty} 1$) as in \cite{pn_journal}. In practice, one can either optimize $K_i$ (from Proposition 1) or $u$, depending on which parameter is easier to adjust. While optimizing $K_i$ offers more control as there are $N_T$ parameters to adjust, finding the optimal set may be challenging. In contrast, optimizing $u$ provides less control but is more straightforward as it is the same for all agents. Similar to optimizing \( K_i \) in Proposition 1, even if some agents have large \( \lambda_i \) and are computationally difficult to optimize, the upper bound can still be controlled by optimizing the worst environments of other agents, unlike the single-agent case \cite{pn_journal}.

\begin{corollary} \normalfont
    Let $\bar{s}$ be an uncoordinated joint state. Then, the joint \( Q \)-functions converge to the sum of the optimal individual \( Q \)-functions of all agents in mean square:
    \begin{align}
        \lim_{t \rightarrow \infty} \bar{Q}_t(\bar{s},\bar{a}) \xrightarrow{\text{in mean square}} \sum_{i=1}^{N_T} Q_i^*(s_i,a_i).
    \end{align}
    \label{Corollary_2}
\end{corollary}

This result demonstrates that, in uncoordinated states, the optimal joint $Q$-functions can be obtained by summing the individual $Q$-functions of all agents. This result relates $Q$-tables of different sizes ($\bar{Q}$: $|\mathcal{S}|$$\times$$|\mathcal{A}|$ vs $\bar{Q}_i$: $|\mathcal{S}_i|$$\times$$|\mathcal{A}_i|$) unlike \cite{pn_journal}, which considers $Q$-tables of the same size.

\subsection{Deterministic Analysis}

We now provide several deterministic results. These results are valid without any assumptions on $Q$-functions such as (\ref{Equ: distribution_assumption}). We adapt the deterministic results and techniques from \cite{ln_journal}, which were originally derived for single-agent MEMQ, to establish our results herein for multi-agent MEMQ.
\begin{proposition} \normalfont
    Let $\bar{s}$ be an uncoordinated joint state. Let the joint $Q$-function update at time $t$ be $\bar{\Delta}_t(\bar{s},\bar{a}) = \bar{Q}_t(\bar{s},\bar{a}) - \bar{Q}_{t-1}(\bar{s},\bar{a})$. Then, the following holds: 
    \begin{align}
        \lim_{t\rightarrow\infty}|\bar{\Delta}_{t}(\bar{s},\bar{a}) - \bar{\Delta}_{t-1}(\bar{s},\bar{a})| = 0.
    \end{align}
\end{proposition}

This proposition provides insight into the behavior of the joint $Q$-function updates over time in uncoordinated states. As more iterations are performed, the updates to the joint $Q$-function become increasingly stable, indicating the stabilization of the overall learning process. This result applies to joint $Q$-functions and thus provides generalization over to the single-agent case \cite{ln_journal}, which is restricted to individual $Q$-functions. We leverage the stability of each agent from \cite{ln_journal} to establish joint stability, since in uncoordinated states agents act independently and the single-agent MEMQ results apply directly to each agent.

\begin{corollary} \normalfont
    Let $\bar{s}$ be an uncoordinated joint state. Then, it requires at most $t_\beta$ iterations to ensure that $|\bar{\Delta}_{t}(\bar{s},\bar{a})|\leq\beta$ for any $\beta>0$, where:
    \begin{align}
        t = \frac{\log(1 - \frac{\beta}{\sum_{i=1}^{N_T}\sum_{n=1}^{K_i}\theta_i^{(n)}(s_i,a_i)})}{\log(u)},
    \end{align}
and $u$ is the constant update ratio, $\theta^{(n)}_i(s_i,a_i)$ is the smallest constant satisfying $|\epsilon^{(n)}_{i,t}(s_i,a_i)| \leq \theta_i^{(n)}(s_i,a_i)$ for all $t$, and $\epsilon_{i,t}^{(n)}(s_i,a_i) = w_{i,t}^{(n)} Q_{i,t}^{(n)}(s_i,a_i) - w_{i,t-1}^{(n)} Q_{i,t-1}^{(n)}(s_i,a_i)$ is the weighted $Q$-function update of $n^{th}$ environment for agent $i$.
\end{corollary}

This result is valid for any $t$; thus, it can be used as a stopping condition for Algorithm \ref{Algorithm: multi_agent_memq}. Moreover, by appropriately choosing the parameters $u$ and $K_i$, we can effectively control the speed of convergence and the accuracy of the $Q$-function updates. This result reflects the sample complexity of Algorithm \ref{Algorithm: multi_agent_memq} to achieve $\beta$-convergence, which is different than the standard sample complexity, which considers $\bar{Q}_t - \bar{Q}^*$ yet we herein consider $\bar{Q}_t - \bar{Q}_{t-1}$. This result considers aggregated performance of all agents and requires more iterations to match the sample complexity of the single-agent case \cite{pn_journal}.

\begin{proposition} \normalfont
    Let $\bar{s}$ be an uncoordinated joint state. If \( |\epsilon^{(n)}_{i,t-1}(s_i,a_i)| \geq |\epsilon^{(n)}_{i,t}(s_i,a_i)| \) for all \( n, t, i \), then:
    \begin{align}
        \lim_{t\rightarrow\infty} |\bar{\Delta}_{t}(\bar{s}, \bar{a})| = 0.
    \end{align}
\end{proposition}

This proposition provides a sufficient deterministic condition for the convergence of Algorithm 1 in uncoordinated states. Unlike Corollary \ref{Corollary_2}, which assumes zero-mean errors in \( Q \)-functions, this result only requires the monotonicity of the weighted \( Q \)-function updates.

\subsection{Analysis of State Misdetection Probability}

In real-world wireless networks, RSS measurements are noisy due to several factors, including interference from nearby transmitters, multipath propagation, or environmental factors such as physical obstacles \cite{molisch2012wireless}. To this end, we employ the noisy ARSS measurement model as follows:
\begin{align}
    I_{i,t} = I_{i,t,\text{true}} + n_t, \label{Equ: noisy_arss}
\end{align}
where $I_{i,t}$ and $I_{i,t,\text{true}}$ are the noisy and true ARSS measurement by TX$_i$ due to the other transmitters at time $t$ and $n_t$ is additive zero-mean Gaussian noise as:
$n_t \sim N(0,\sigma_u^2)$ in uncoordinated states and $n_t \sim N(0,\sigma_c^2)$ in coordinated states. In uncoordinated states, transmitters are often far apart from each other, leading to higher measurement uncertainty due to a lack of information exchange. Conversely, in coordinated states, closer proximity and synchronization reduce this uncertainty. Hence, we assume $\sigma_u \gg \sigma_c$. Additionally, we define the probability of state misdetection as:
\begin{align}
    P_{\mathrm{mis}} &= P(\text{detect U} \mid \text{actual C}) P(\text{actual C}) \nonumber \\
    &\quad + P(\text{detect C} \mid \text{actual U}) P(\text{actual U}),
\end{align}
where \( P(\text{detect U} \mid \text{actual C}) \) denotes the probability that a joint state actually requires coordination but is incorrectly classified by the agents as an uncoordinated state.

Using the noisy ARSS model (17), we can derive a lower bound on \( P_{\mathrm{mis}} \) as a function of \( I_{\mathrm{thr}} \) (see the supplementary material), which represents the \textbf{unavoidable} state misdetection error. Hence, varying \( I_{\mathrm{thr}} \) directly impacts the accuracy of state detection. If \( I_{\mathrm{thr}} \) is set too high, most states will be misclassified as uncoordinated, even when coordination is beneficial. This leads to high policy error due to missed opportunities for joint optimization, but communication and joint updates are rarely triggered, resulting in minimal runtime and overhead. Conversely, setting \( I_{\mathrm{thr}} \) too small causes nearly all states to be classified as coordinated, yielding excessive communication with the leader and frequent joint \( Q \)-function updates even when unnecessary. This may improve policy accuracy, yet incurs higher computational overhead. To this end, we choose \( I_{\mathrm{thr}} \) to minimize the lower bound on \( P_{\mathrm{mis}} \). The optimal \( I_{\mathrm{thr}} \) admits a closed-form expression as given in the supplementary material, and yields the following bounds on \( P_{\mathrm{mis}} \).
\begin{proposition} \normalfont
\label{prop4_multi_memq}
    Assume there are $N_T = 2$ transmitters, and $|\mathcal{S}_C|\sigma_u > |\mathcal{S}_U|\sigma_c$. Under the noisy ARSS model (\ref{Equ: noisy_arss}), the probability of misdetecting the state as coordinated or uncoordinated is bounded as:
    \begin{align}
    P_{\text{mis}} \in \Bigg[\mathcal{Q}\left(\frac{\Delta_I}{\sigma_c}\right) &\frac{|\mathcal{S}_C|}{|\mathcal{S}|} + \mathcal{Q}\left(\frac{-\Delta_I}{\sigma_u}\right) \frac{|\mathcal{S}_U|}{|\mathcal{S}|}, \nonumber\\ &\mathcal{Q}\left(\frac{-\Delta_I}{\sigma_c}\right) \frac{|\mathcal{S}_C|}{|\mathcal{S}|} + \mathcal{Q}\left(\frac{\Delta_I}{\sigma_u}\right) \frac{|\mathcal{S}_U|}{|\mathcal{S}|}\Bigg],
\end{align}
where $\mathcal{Q}(z) = \frac{1}{2} \operatorname{erfc}\left(\frac{z}{\sqrt{2}}\right)$ and 
\begin{align}
    \Delta_I = \sqrt{2\ln\left(\frac{|\mathcal{S}_C| \sigma_u}{|\mathcal{S}_U|\sigma_c}\right)\left(\frac{1}{\sigma_c^2} - \frac{1}{\sigma_u^2}\right)^{-1}}.
\end{align}
\end{proposition}

The lower bound represents the fundamental limit on detection accuracy due to noise in ARSS measurements, reflecting an unavoidable misdetection error. The upper bound indicates the maximum misdetection probability, representing the worst-case scenario where the system is most vulnerable to errors from noisy ARSS measurements. We note that the lower and upper bounds are complementary and sum to 1. Thus, an increase in the lower bound directly results in a decrease in upper bound, and vice versa. For example, as $\sigma_c$ increases, the lower bound increases, indicating a higher minimum error probability. On the other hand, the upper bound decreases because higher noise in coordinated states causes the ARSS measurements to resemble those of uncoordinated states. As a result, the likelihood of misdetecting all states in the worst-case scenario decreases. When \( |\mathcal{S}_C| \to |\mathcal{S}_U| \) and \( \sigma_C \to \sigma_U \), both bounds converge to 0.5, indicating that the system can no longer distinguish between coordinated and uncoordinated states, leading to random detection.

\begin{corollary} \normalfont
The bounds on the probability of misdetection can be generalized to any number of agents $N_T > 2$ using Taylor series approximation as follows:
\begin{align}
    P_{mis} \in \Bigg[& \Big(1-\sum_{i=1}^{N_T} \mathcal{Q}\Big(\frac{I^l_{thr} - I_{i,t,true}}{\sigma_c}\Big)\Big)\frac{|\mathcal{S}_c|}{|\mathcal{S}|} + \nonumber\\
    & \hspace{40pt}\mathcal{Q}\Big(\frac{I^l_{thr} - I_{i',t,true}}{\sigma_u}\Big)\frac{|\mathcal{S}_u|}{|\mathcal{S}|}, \nonumber\\
    & \sum_{i=1}^{N_T} \mathcal{Q}\Big(\frac{I^{u}_{thr} - I_{i,t,true}}{\sigma_u}\Big)\frac{|\mathcal{S}_u|}{|\mathcal{S}|} + \nonumber\\
    & \hspace{40pt}\Big(1 - \mathcal{Q}\Big(\frac{I^{u}_{thr} - I_{i',t,true}}{\sigma_c}\Big)\Big)\frac{|\mathcal{S}_c|}{|\mathcal{S}|}\Bigg],
\end{align}
where \( I^l_{thr} \) and \( I^{u}_{thr} \) are two distinct constants as a (non-linear) function of \( N_T \), \( |\mathcal{S}_c| \), \( |\mathcal{S}_u| \), \( \sigma_c \), \( \sigma_u \) and the minimum true ARSS measurement among all agents $I_{i',t,true} = \underset{i=1,\dots,N_T}{\min} \, I_{i,t,true}$.
\end{corollary}

This result is derived using a second-order Taylor series approximation of the Gaussian \( \mathcal{Q}(x) \) function around \( x = 0 \) and hence is numerically accurate when the threshold \( I_{thr} \) is close to the true ARSS measurements \( I_{i,t,true} \) or when noise variances \( \sigma_c \) and \( \sigma_u \) are large. It is possible to employ a higher-order Taylor approximation for a more precise bound, yet it becomes harder to derive a closed-form expression. The lower and upper bounds herein are complementary (i.e., one increases while the other one decreases) as in the two-agent case. We will numerically analyze the effect of individual parameters on the tightness of the bound.

Proposition 4 and Corollary 4 present novel results compared to the single-agent case \cite{pn_journal, ln_journal}. These results are derived for the wireless network in Sec.III using the ARSS model (\ref{Equ: noisy_arss}); however, similar results can be derived for the multi-agent versions of MISO and MIMO networks in \cite{ln_journal} with different ARSS models such as multiplicative noise model e.g. $I_{i,t} = h_t I_{i,t,\text{true}} + n_t$, with $h_t$ being the fading coefficient or interference noise model $I_{i,t} = I_{i,t,\text{true}} + \sum_{j \neq i} I_{j,t,\text{interference}} + n_t$.

\section{Numerical Results}\label{sec:numerical_results}

In this section, we consider a variety of performance metrics to assess the accuracy and complexity performance of Algorithm \ref{Algorithm: multi_agent_memq}. The following parameters are used: \( T = 10^5 \), \( l = 30 \), \( \gamma = 0.95 \), 
\( \alpha_t = \frac{1}{1 + \sfrac{t}{1000}} \), \( \zeta_t = \max(0.99^t, 0.01) \), \( u_t = 1 - e^{\sfrac{-t}{1000}} \) (or $u = 0.5$ ), \( L = 100 \), \( \Delta_L = 2 \), \( r_j \sim \operatorname{unif}\left[\frac{L}{2}, \frac{3L}{4}\right] \), 
\( \Delta_I = 50 \), \( \sigma = 1 \), \( P_i = 10^{-2} \), 
\( \beta_1 = \beta_2 = \beta_3 = \frac{1}{3} \). The parameters \( I_{max} \) and \( I_{min} \) are set to the max and min ARSS measurements for each simulation. The parameters $\lambda_{i,n}$ are estimated numerically as in \cite{pn_journal}. \( I_{thr}\) is chosen to be the minimizer of the lower bound on \( P_{mis} \) from Proposition 4 and Corollary 4. \( K_i \) is selected proportional to the state-space size as in \cite{pn_journal}, and remains constant for all $i$. The values of $\sigma_c$ and $\sigma_u$ are chosen such that $\sigma_u \gg \sigma_c$ and the assumption in Proposition \ref{prop4_multi_memq} is satisfied. Simulations are averaged over 100 independent simulations.

\renewcommand{\arraystretch}{1.3}
\begin{table}[t]
    \scriptsize
    \centering
    \begin{tabular}{|l|l|}
        \hline
        \textbf{Algorithm} & \textbf{Type} 
        \\ \hline
        Randomized Ensembled Double Q-learning (REDQ) \cite{randomized_double_q} & Fully centralized \\ 
        \hline
        Multi-agent Deep Q-Networks (DQN) \cite{ctde_1} & CTDE \\ 
        \hline
        QD-Learning (QD) \cite{networked_agent_1} & DNA  \\ 
        \hline
        Two-layer Q-learning (TQ) \cite{networked_agent_3} & DNA \\
        \hline
        Ideal Independent Q-learning (I2Q) \cite{independent_2} & Fully decentralized \\ 
        \hline
        Hysteretic Q-learning (HQ) \cite{independent_1} & Fully decentralized \\ 
        \hline
    \end{tabular}
    \caption{Different cooperative MARL algorithms}
    \label{Table:multi_agent_Q_learning}
\end{table}

\subsection{Policy and $Q$-function accuracy}\label{subsec: APE_results}

We first define the \textit{average policy error (APE)} as:
\begin{align}
    \text{APE} = \frac{1}{|\mathcal{S}|} \sum_{\bar{s}=1}^{|\mathcal{S}|} \mathbf{1}(\bar{\pi}^{*}(\bar{s}) \neq \bar{\pi}(\bar{s}))\label{Equ:APE}
\end{align}
where $\bar{\pi}^{*}$ is the optimal joint output policy from (\ref{Equ:multi_agent_minimization}), and $\bar{\pi}$ is the joint output policy of Algorithm \ref{Algorithm: multi_agent_memq}. We also define the \textit{average $Q$-function distance (AQD)} as:
\begin{align}
    \text{AQD} = \frac{1}{|\mathcal{S}||\mathcal{A}|} \sum_{\bar{s},\bar{a}} (\bar{Q}^{*}(\bar{s},\bar{a})-\bar{Q}(\bar{s},\bar{a}))^2 \label{Equ:joint_AQD}
\end{align}
where $\bar{Q}^{*}$ is the optimal joint $Q$-functions, and $\bar{Q}$ is the joint output $Q$-functions of Algorithm \ref{Algorithm: multi_agent_memq} at time $t$. We evaluate our algorithm against several fully cooperative tabular and deep MARL algorithms as shown in Table \ref{Table:multi_agent_Q_learning}. $\bar{Q}^\star$ and $\bar{\pi}^\star$ are computed via centralized $Q$-learning by treating all agents as a single composite agent with full joint information. This is used \textbf{only} for benchmarking and \textbf{not} required in practice.

\begin{figure}[t]
    \centering
    \subfloat[APE vs ($N_T, N_B$) \label{fig:ape_vsNbNt}]{{\includegraphics[width=0.45\textwidth]{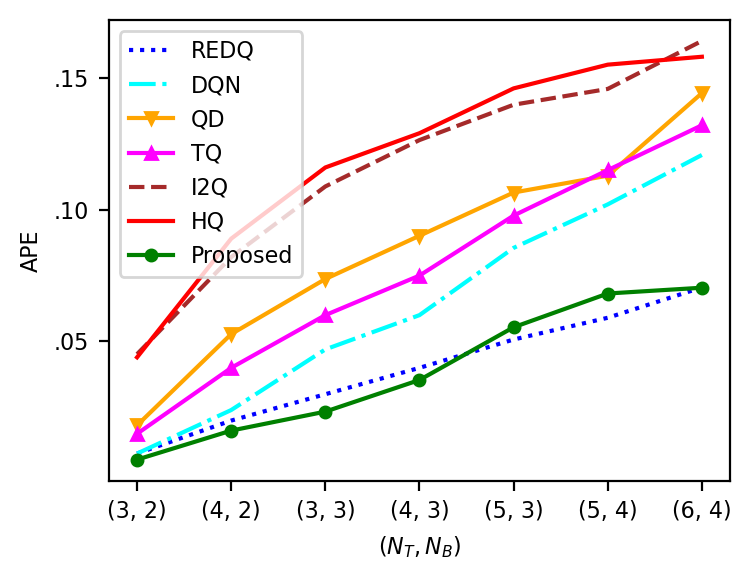}}}
    
    \subfloat[AQD vs iterations \label{fig:Qfunc_diff}]
    {{\includegraphics[width=0.45\textwidth]{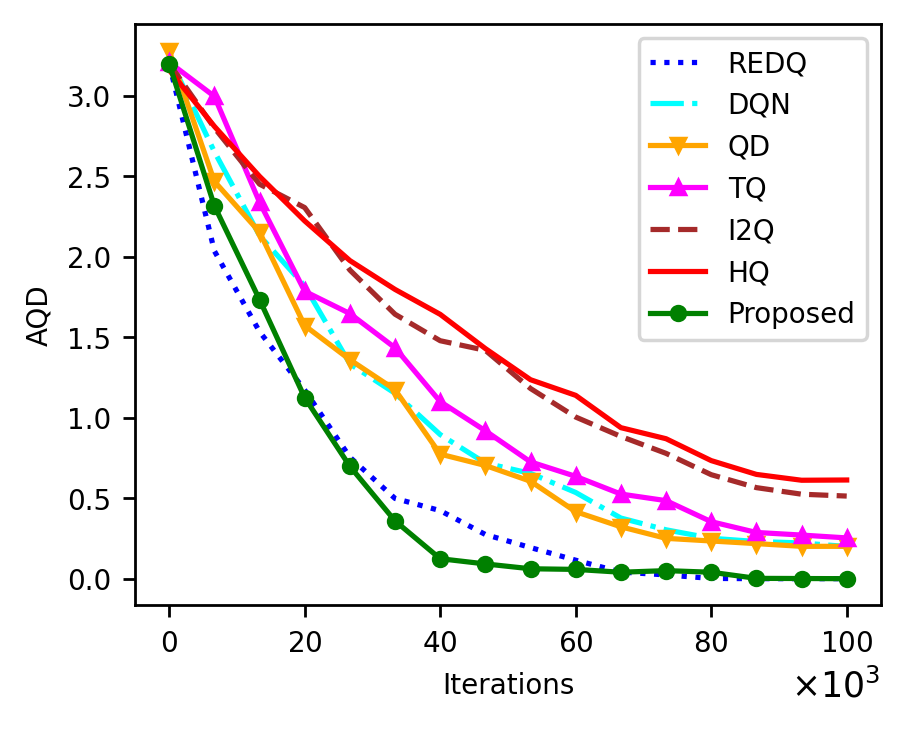}}}
    \caption{Performance of different MARL algorithms}
\end{figure}

APE results across an increasing number of transmitters and base stations $(N_T, N_B)$ are shown in Fig.\ref{fig:ape_vsNbNt}. Our algorithm offers 40\% less APE than CTDE, 45\% less APE than DNA, and 60\% less APE than fully decentralized algorithms, particularly across large $(N_T, N_B)$. These APE gains can be attributed to the following reasons: (i) exchanging local ARSS information through the leader can effectively capture coordination among agents. (ii) Leveraging synthetic environments through MEMQ helps agents improve their exploration capabilities \cite{pn_journal}. (iii) MEMQ exploits structural similarities among multiple environments, providing an advantage over neural network-based algorithms that do not utilize such properties. (iv) Optimizing $I_{thr}$, by minimizing the state misdetection probability, achieves the optimal $|\mathcal{S}_c| / |\mathcal{S}_u|$ ratio, which combines coordination benefits in coordinated states and complexity gains in uncoordinated states. The centralized algorithm achieves slightly better APE than our algorithm, as expected since it assumes full observability among agents. However, the performance difference is minimal (only less than 10\%) because (i) communication and information exchange among agents only in coordinated states are sufficient to capture coordination, and (ii) periodically resetting the belief state prevents error accumulation and yields accurate joint state estimates. We also note that APE improvements increase as the wireless network grows, demonstrating the scalability of the proposed algorithm.

Our algorithm also supports highly heterogeneous users. 
We illustrate this with a 6-user system where each user’s parameters are randomly drawn from these sets: mobility levels \(\{1\Delta_L, 2\Delta_L, 3\Delta_L\}\) steps per time slot; number of antennas \(\{1, 2, 3, 4\}\); transmission powers \(\{0.1\,\text{W}, 0.25\,\text{W}, 0.5\,\text{W}\}\); and different coverage areas with radius $\{50\Delta_L, 75\Delta_L, 100\Delta_L\}$. Numerical results show that our algorithm reduces APE by 25\% compared to CTDE, 40\% compared to DNA, 45\% compared to fully decentralized algorithms.

Fig.\ref{fig:Qfunc_diff} shows the AQD results over iterations with $(N_T, N_B) = (5,3)$. $Q$-functions of all algorithms are initialized to very small random but identical values (thus AQD of all algorithms start at the same value). AQD values for all algorithms decrease over time with varying decay rates, convergence behavior, and convergence points. This result demonstrates that our algorithm converges numerically to the optimal $Q$-functions, with a convergence rate 30\% faster than CTDE and DNA algorithms, and 40\% faster fully decentralized algorithms. This result follows from the fast exploration and convergence properties of the single-agent MEMQ and the purely decentralized nature of agents in uncoordinated states. This result also shows the accuracy and convergence of the Bayesian joint state estimation technique. Our algorithm has a similar converge speed to the centralized algorithm, which is consistent with Fig.\ref{fig:ape_vsNbNt}. We also observe that decentralized algorithms (e.g., HQ and I2Q) and deep RL algorithms (e.g., DQN) may not always converge to the optimal $Q$-functions. We underline that algorithms have similar trends in Fig.\ref{fig:ape_vsNbNt} and Fig.\ref{fig:Qfunc_diff} (e.g. HQ has the largest AQD as it does not converge, which also leads to a very large APE in the limit).

We finally evaluate the sensitivity of Algorithm \ref{Algorithm: multi_agent_memq} by varying several wireless parameters within reasonable ranges and measuring the mean change in APE and AQD. Only one parameter is changed each time, while the others are held constant at the values that minimize APE. The results are summarized in Table \ref{tab:sensitivity_assessment}. We observe that APE and AQD vary at most by 8\% and 12\%, which indicates a stable performance. The parameters $L$ and $\Delta_L$ directly affect the state-space size, but the change in these parameters results in a minimal performance change, showing the efficiency of our algorithm in moderately large discrete state-spaces.

\renewcommand{\arraystretch}{1.3}
\begin{table}[t]
    \centering
    \begin{tabular}{|c|c|c|c|}
        \hline
        \textbf{Parameter} & \textbf{Range} & $\Delta_\text{APE}$ & $\Delta_\text{AQD}$\\
        \hline
        $L$ & \{40, 60, ... 400\} & 5\% & 6\%\\
        \hline
        $\Delta_L$ & \{1, 2, ... 10\} & 6\% & 10\%\\
        \hline
        $r_j$ & $\operatorname{unif}[x,3x], x\in\{\frac{L}{5}$,$\frac{L}{4}$,$\frac{L}{3}$,$\frac{L}{2}$\} & 8\% & 8\%\\
        \hline
        $\Delta_I$ & \{10, 20, ... 100\} & 8\% & 12\%\\
        \hline
        $\sigma$ & \{0.2, 0.4, ... 2\} & 6\% & 10\%\\
        \hline
        $P_i$ & \{$10^{-4}$,$10^{-3}$,$10^{-2}$,$10^{-1}$,$1$\} & 6\% & 6\%\\
        \hline
    \end{tabular}
    \caption{Impact of wireless parameters in performance.}
    \label{tab:sensitivity_assessment}
\end{table}

\subsection{Complexity}\label{subsec: runtime_results}

\begin{figure}[t]
    \scriptsize
    \centering
    \includegraphics[width=0.45\textwidth]{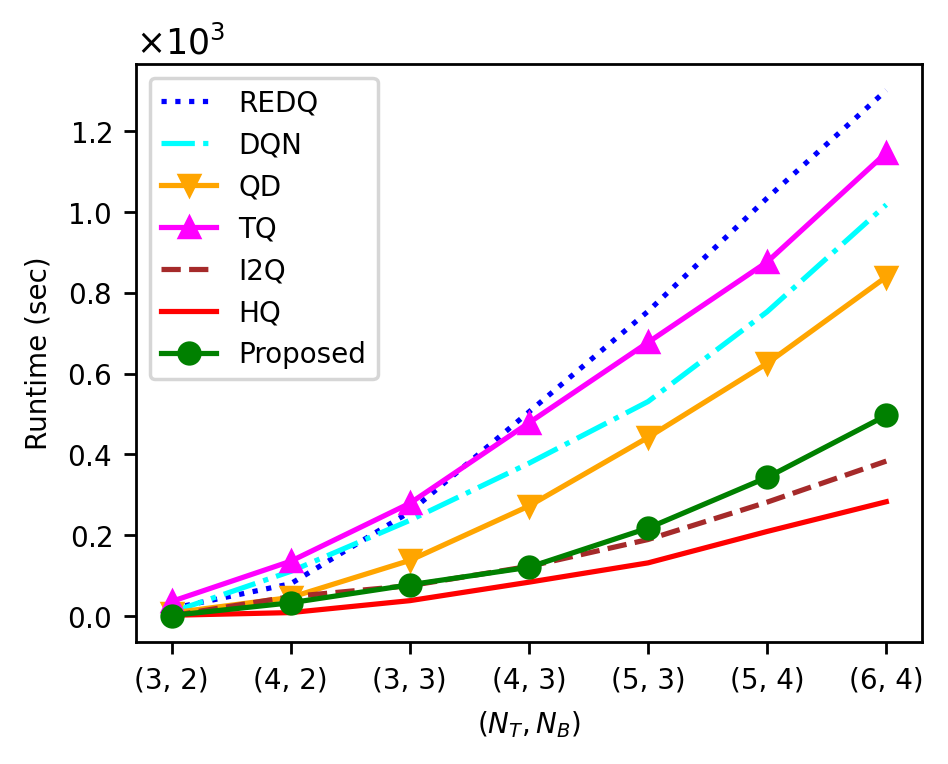}
    \caption{Runtime of different MARL algorithms}
    \label{fig:runtime_Q}
\end{figure}

In order to evaluate the complexity of Algorithm \ref{Algorithm: multi_agent_memq}, we first consider \textit{computational runtime complexity} -- the time required for convergence with optimized parameters on the same hardware. Though runtime varies by hardware, we focus on the relative complexity between different algorithms. Some algorithms may not converge to the optimal $Q$-functions; in such cases, we terminate the algorithm once it converges to a stable point. Fig.\ref{fig:runtime_Q} shows the runtime results across increasing $(N_T,N_B)$. While all algorithms show an increase in runtime as the wireless network grows, our algorithm has a relatively small increase. In particular, our algorithm can achieve up to 45\% less runtime than fully centralized, CTDE and DNA algorithms, particularly across large $(N_T,N_B)$. This efficiency stems from (i) a fully decentralized approach in uncoordinated states, which eliminates unnecessary coordination and information exchange among agents; (ii) minimal information sharing in coordinated states through only the leader transmitter, which avoids expensive and redundant communication between all agents, and (iii) leveraging fast exploration capabilities of digital cousins and fast convergence of single-agent MEMQ. While decentralized algorithms assume full independence between agents without any coordination or communication, and hence have the smallest runtime, our algorithm has only 10\% larger runtime than the decentralized algorithms. We underline that this result is consistent with Fig.\ref{fig:Qfunc_diff} as plotting the AQD values of algorithms over iterations for different $(N_T, N_B)$ values yields similar plots to Fig.\ref{fig:runtime_Q} (i.e. similar algorithm orders).

One can also consider the \textit{algorithm sample complexity} -- the number of training samples (or interactions with the environment) needed to achieve good policies. To this end, we can determine the time index at which the AQD in (\ref{Equ:joint_AQD}) falls below some threshold. This index is proportional to the sample complexity, as the total number of training samples obtained by iteration $t$ is $t$ x $l$. Fig. \ref{fig:Qfunc_diff} provides an example where our algorithm achieves an AQD of less than 1 in 24000 iterations, while other algorithms need at least 40000 iterations, resulting in a 40\% reduction in sample complexity. This improvement becomes greater for smaller thresholds (e.g. 50\% reduction in sample complexity to reach AQD < 0.5).

Extensive simulations also show that both the computational complexity of our algorithm are primarily dominated by the slowest agent (i.e. the slowest MEMQ). This is expected, as the joint system cannot converge until all individual systems do. Thus, optimizing the slowest agent is an effective strategy for enhancing the overall complexity of the joint system. The memory complexity of Algorithm \ref{Algorithm: multi_agent_memq} (storing local and global $Q$-functions for each agent), is not explicitly addressed in this paper; however, it can be handled by employing the structural state-aggregation algorithm which is presented in \cite{talha_eusipco} and successfully employed for wireless networks in \cite{ln_journal}. 

\subsection{Numerical consistency of propositions}

\begin{figure}[!t]
    \centering
    \subfloat[Simulation variance vs. upper bounds on variance for uncoordinated states \label{Fig:simulation_variance_unc}]
    {{\includegraphics[width=0.4\textwidth]{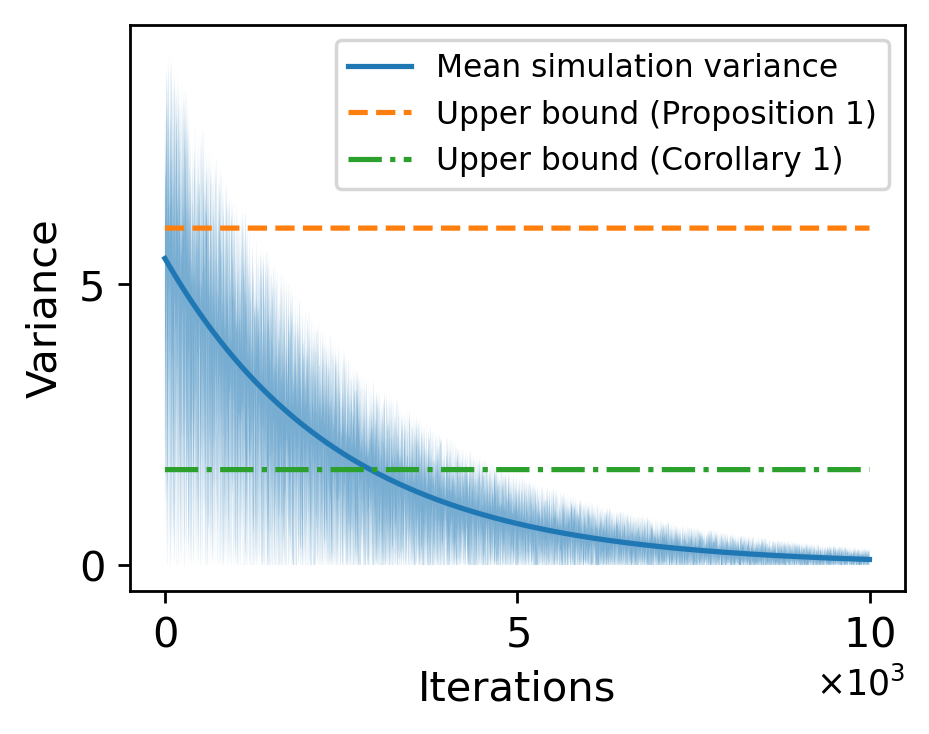}}}
    \vspace{2pt}
    \subfloat[Simulation variance vs. upper bounds on variance for coordinated states\label{Fig:simulation_variance_coor} ]{{\includegraphics[width=0.4\textwidth]{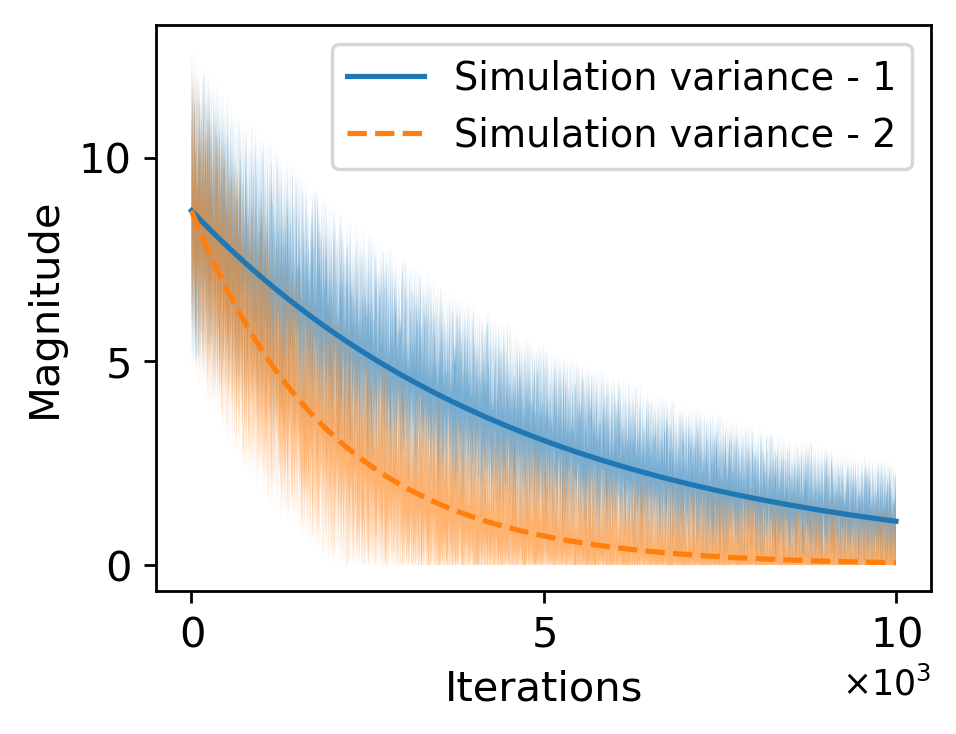}}}
    \vspace{2pt}
    \subfloat[Behavior of $\bar{\Delta}_{t}(\bar{s},\bar{a})$\label{Fig:delta_magnitude}]
    {{\includegraphics[width=0.4\textwidth]{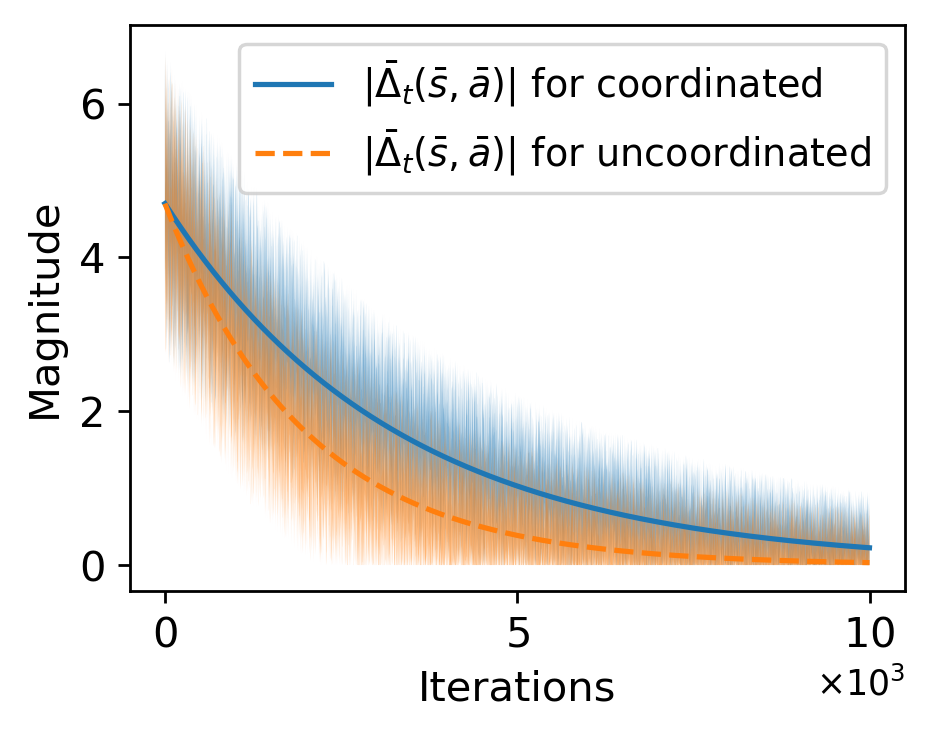}}}
    \caption{Numerical validation of theoretical results.}
\end{figure}

\begin{figure}[!t]
    \centering
    \subfloat[Simulation $P_{mis}$ vs. upper and lower bounds on $P_{mis}$ with $N_T = 2$ \label{Fig:p_mis_error_1}]
    {{\includegraphics[width=0.4\textwidth]{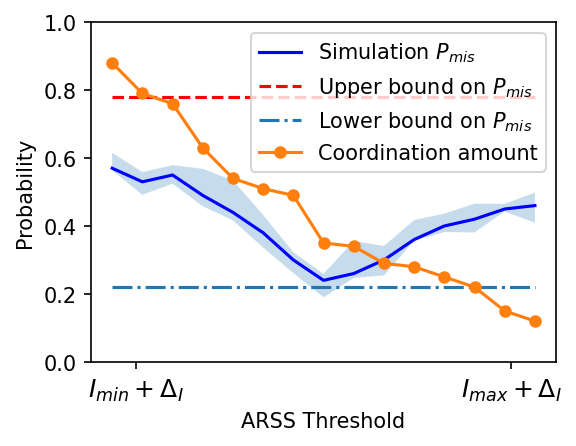}}}
    \vspace{2pt}
    \subfloat[Simulation $P_{mis}$ vs. upper and lower bounds on $P_{mis}$ with $N_T = 5$  \label{Fig:p_mis_error_2}]
    {{\includegraphics[width=0.4\textwidth]{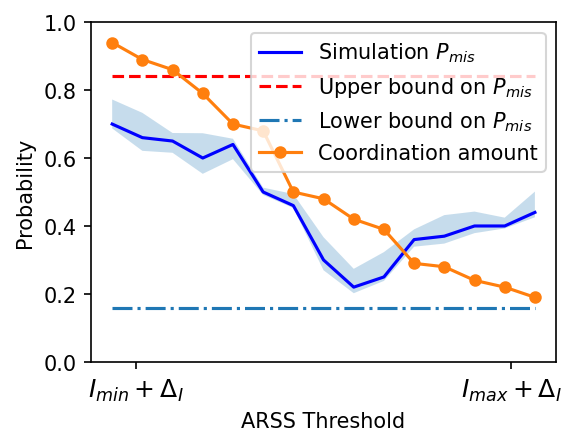}}}
    \caption{Numerical validation of theoretical results.}
\end{figure}

In this section, we simulate the results in the propositions and compare theoretical results with simulation results. The same numerical parameters in Section VI are employed (with $N_T$ = 3, $N_B$ = 2). We first numerically compute the variance expression in Proposition 1 as in \cite{pn_journal}:
\begin{align}
    {\mathbb{V}[\mathcal{E}_t(\bar{s},\bar{a})] \approx \frac{1}{2\Delta_t}\sum_{i = t - \Delta_t}^{t + \Delta_t} \mathcal{E}_i(\bar{s},\bar{a})^2 \text{--} \Big[\frac{1}{2\Delta_t}\sum_{i = t - \Delta_t}^{t + \Delta_t} \mathcal{E}_i(\bar{s},\bar{a})\Big]^2},
\end{align}
where $\mathcal{E}_t(\bar{s},\bar{a}) = \bar{Q}_t(\bar{s},\bar{a}) - \sum_{i=1}^{N_T} Q_i^*(s_i,a_i)$ with $\Delta_t \ll t$. 

The upper bounds on the variance from Proposition 1 and Corollary 1 (with $K_1$ = $K_2$ = $K_3$ = 2) and the simulation variance for the \textbf{uncoordinated} state-action pair $(\bar{s}, \bar{a}) = (10, 2)$ are shown in Fig.\ref{Fig:simulation_variance_unc}. The mean error variance approaches 0 with more iterations. The upper bound from Proposition 1 holds for all $t$, whereas the upper bound from Corollary 1 holds in the limit (i.e. when the algorithm converges). This result also validates the assumption (\ref{Equ: distribution_assumption}) as both bounds hold. A similar plot for the expectation of $\mathcal{E}(\bar{s}, \bar{a})$ can also be shown, which numerically validates the mean-square convergence of Algorithm \ref{Algorithm: multi_agent_memq} in uncoordinated states.

Fig.\ref{Fig:simulation_variance_coor} shows two variances for the \textbf{coordinated} state-action pair $(\bar{s}, \bar{a}) = (40, 1)$. The orange dotted curve represents the variance of $\bar{Q}_t(\bar{s}, \bar{a}) - \sum_{i=1}^{N_T} Q_i^*(s_i, a_i)$ (which converges to 0) while the blue solid curve shows the variance of $\bar{Q}_t(\bar{s}, \bar{a}) - \bar{Q}^*(\bar{s}, \bar{a})$ (which converges to some non-zero value), and the shaded areas represent the standard deviation over 200 simulations. This result shows that the joint $Q$-functions of coordinated states converge to $\bar{Q}^*(\bar{s}, \bar{a})$, not to the sum of individual optimal $Q$-functions $\sum_{i=1}^{N_T} Q_i^*(s_i, a_i)$ as in uncoordinated states, due to coordination between agents, which also shows that $\bar{Q}^*(\bar{s}, \bar{a}) \neq \sum_{i=1}^{N_T} Q_i^*(s_i, a_i)$. Although these two quantities are related, deriving a closed-form relationship is challenging as it depends on real-time ARSS measurements and the ARSS threshold.

Fig.\ref{Fig:delta_magnitude} shows the evolution of $|\bar{\Delta}_{t}(\bar{s}, \bar{a})|$ over time for the \textbf{coordinated} state-action pair $(\bar{s}, \bar{a}) = (40, 1)$ and \textbf{uncoordinated} state-action pair $(\bar{s}, \bar{a}) = (10, 2)$. Clearly, it approaches 0 (with different speeds) for both state-action pairs, which verifies the deterministic convergence of Algorithm 1, and the validity of Proposition 3 (and thus Proposition 2). We underline that while Propositions 2 and 3 are derived for uncoordinated states, the results apply to \textbf{both} uncoordinated and coordinated states.

Fig.\ref{Fig:p_mis_error_1} illustrates the simulation misdetection probability ($P_{mis}$) vs. ARSS threshold $I_{thr}$, which varies in [$I_{min}, I_{max}$], along with the upper and lower bounds from Proposition 4 for $N_T = 2$, and the coordination percentage (i.e. $\frac{|\mathcal{S}_c|}{|\mathcal{S}|}$), where the blue shaded area represents the standard deviation over 200 simulations. We make several observations. First, the upper and lower bounds in Proposition 4 hold for any ARSS threshold value. Second, there is a unique ARSS threshold that minimizes the $P_{mis}$, which can also be theoretically obtained in Proposition 4. Third, as the ARSS threshold increases, the proportion of coordinated states decreases monotonically. When $P_{mis}$ is minimized, coordinated states make up around 35\%, indicating that even in the optimal scenario, the system largely operates independently. Fourth, $P_{mis}$ is larger when $I_{thr}$ is very small (close to the minimum ARSS measurement) than when it is very large (close to the maximum ARSS measurement). This is expected as large $I_{thr}$ causes most states to be uncoordinated, reducing the likelihood of misdetection. Similarly, even though $I_{thr}$ changes significantly, the change in $P_{mis}$ remains relatively modest, demonstrating the robustness of the system to change in system parameters as in Table \ref{tab:sensitivity_assessment}.

Fig.\ref{Fig:p_mis_error_2} presents very similar results for the \( N_T = 5 \) case along with the bounds from Corollary 2. Similar results also hold for larger $N_T$.) Both bounds hold across all \( I_{thr} \) values, but are relatively looser compared to \( N_T = 2 \) case. The variation in \( P_{mis} \) is also larger, indicating that the system is more sensitive to the threshold choice -- this is expected as coordination now depends on inputs from multiple agents. When \( I_{thr} \) is very small, most states are coordinated since more agents in close proximity lead to higher ARSS values, which will likely be larger than the ARSS threshold, making the states coordinated. Similar trends can be observed for larger $N_T$ with relatively larger $P_{mis}$ for a given \( I_{thr} \) and looser bounds.

\section{Conclusions}\label{sec:conclusions}

We present a novel multi-agent MEMQ algorithm for cooperative decentralized wireless networks with multiple networked transmitters (TXs) and base stations (BSs) to address the performance and complexity limitations of single-agent MEMQ and existing MARL algorithms. Transmitters lack access to global information but can locally measure ARSS due to other TXs. TXs act independently to minimize their individual costs in uncoordinated states. In coordinated states, they employ a Bayesian approach to estimate the joint state based on local ARSS measurements and share limited information with only the leader TX to minimize the joint cost. This communication scheme incurs an information-sharing cost that scales only linearly with the number of TXs, independent of the joint state-action space size. We provide several deterministic and probabilistic theoretical results, including convergence, upper bounds on estimation error variance, and state misdetection probability. We numerically show that our algorithm outperforms several decentralized and CTDE algorithms by achieving 60\% lower APE with 45\% lower runtime complexity and 40\% faster convergence with 40\% less sample complexity. It significantly reduces the complexity of centralized algorithms while achieving comparable APE. It is shown that the simulation results closely follow the theoretical results. For our future work, we work on developing a theoretical sample complexity expression to enable direct comparison with MARL algorithms. Our algorithm is not directly applicable to continuous state–action spaces. To this end, we are extending the tabular multi-agent MEMQ framework to deep RL using an actor–critic architecture. Finally, we focus on devising efficient sampling methods to ensure good data coverage across all environments and all agents as in \cite{talha_spawc, talha_coverage_journal}.

\bibliographystyle{IEEEtran}
\bibliography{references.bib}

\end{document}